\documentclass{article} 
\usepackage[final]{colm2026_conference}

\usepackage{microtype}
\usepackage{hyperref}
\usepackage{url}
\usepackage{booktabs}
\usepackage{multirow}
\usepackage[table]{xcolor}
\usepackage{textcomp}
\usepackage{graphicx}
\usepackage{amssymb}

\usepackage{booktabs}
\usepackage{multirow}
\usepackage{makecell}
\usepackage[table]{xcolor}
\usepackage{threeparttable}

\usepackage{microtype}
\usepackage{hyperref}
\usepackage{url}
\usepackage{booktabs}

\usepackage{lineno}

\usepackage{graphicx}
\usepackage{url}
\usepackage{ulem}
\usepackage{enumitem}
\usepackage{multirow}
\usepackage{soul}
\usepackage{xcolor}
\usepackage{subcaption}
\usepackage{amsmath}
\usepackage{amssymb}
\usepackage{mathtools}
\usepackage{amsthm}
\usepackage{caption}
\usepackage{wrapfig}
\usepackage{booktabs}
\usepackage{siunitx}
\usepackage{booktabs}
\usepackage{longtable}
\usepackage{tabularx}
\usepackage{soul}
\usepackage{booktabs}
\usepackage{multirow}
\usepackage{siunitx}
\usepackage{setspace}

\definecolor{siggreen}{RGB}{0, 100, 0}     

\definecolor{nomgreen}{RGB}{0, 153, 0}  

\definecolor{nomred}{RGB}{230, 0, 0}      

\newcommand{\best}[1]{\textbf{#1}}
\newcommand{\second}[1]{\underline{#1}}

\newcommand{\sigup}{\rlap{\,\textcolor{siggreen}{$\uparrow\!\uparrow$}}}
\newcommand{\nomup}{\rlap{\,\textcolor{nomgreen}{$\uparrow$}}}
\newcommand{\nomdown}{\rlap{\,\textcolor{nomred}{$\downarrow$}}}

\newcommand{\textsigup}{\textcolor{siggreen}{$\uparrow\!\uparrow$}}
\newcommand{\textnomup}{\textcolor{nomgreen}{$\uparrow$}}
\newcommand{\textnomdown}{\textcolor{nomred}{$\downarrow$}}

\usepackage{algorithm2e, amsmath}
\usepackage{amsthm}

\theoremstyle{definition}
\newtheorem*{definition}{Definition}
\newcommand{\phn}{\phantom{0}}

\usepackage{lineno}

\definecolor{darkblue}{rgb}{0, 0, 0.5}
\hypersetup{colorlinks=true, citecolor=darkblue, linkcolor=darkblue, urlcolor=darkblue}

\title{CT Open: An Open-Access, Uncontaminated, Live Platform \\for the Open Challenge of Clinical Trial Outcome Prediction}

\author{Jianyou Wang$^1$\thanks{equal contributions}, Youze Zheng$^1$\footnotemark[1], Longtian Bao$^1$\footnotemark[1], Hanyuan Zhang$^1$\footnotemark[1],\\
\textbf{Qirui Zheng$^1$, Yuhan Chen$^1$, Yang Zhang$^1$, Matthew Feng$^1$,}\\ 
\textbf{Maxim Khan$^3$, Aditya K. Sehgal$^3$, Christopher D. Rosin$^3$,}\\ \textbf{Ramamohan Paturi$^1$, Umber Dube$^2$, Leon Bergen$^1$} \\
$^1$Laboratory for Emerging Intelligence, University of California San Diego \\
$^2$Department of Dermatology, University of California San Diego \\
$^3$Elsevier \\
\texttt{\{jiw101, yoz018, haz146\}@ucsd.edu}}

\begin{document}

\ifcolmsubmission
\linenumbers
\fi

\maketitle

\begin{abstract}
Scientists have long sought to accurately predict outcomes of real-world events before they happen. Can AI systems do so more reliably? We study this question through clinical trial outcome prediction, a high-stakes open challenge even for domain experts. We introduce \textbf{CT Open}, an open-access, live platform that would run four challenge every year. Anyone can submit predictions for each challenge. CT Open evaluates those submissions on trials whose outcomes were not yet public at the time of submission but became public afterwards. Determining if a trial's outcome is public on the internet before a certain date is surprisingly difficult. Outcomes posted on official registries may lag behind years, while the first mention may appear in obscure articles. To address this, we propose a novel, fully automated decontamination pipeline that uses iterative LLM-powered web search to identify the earliest mention of trial outcomes. We validate the pipeline’s quality and accuracy by human expert's annotations. Since CT Open’s pipeline ensures that every evaluated trial had no publicly reported outcome when the prediction was made, it allows participants to use any methodology and any data source. In this paper, we release a training set and two time-stamped test benchmarks, Winter 2025 and Summer 2025. We believe CT Open can serve as a central hub for advancing AI research on forecasting real-world outcomes before they occur, while also informing biomedical research and improving clinical trial design. Code to reproduce our experiments is available at \href{https://github.com/ClinicalTrial-OpenChallenge/CT_Open}{https://github.com/ClinicalTrial-OpenChallenge/CT\_Open}.
\end{abstract}

\section{Introduction}

Large language model's performance is increasingly competitive with humans on a wide range of tasks. This progress raises a natural question: beyond assisting humans on tasks they can already perform, can AI systems make reliable predictions on high-stakes problems whose answers are not yet known? In this paper, we study one such problem: predicting the outcome of a clinical trial before the trial has published results or before it has even begun.

Clinical trial outcome prediction is both difficult and consequential. If the problem were easy, clinical development would not remain so uncertain, with many launched trials failing their objectives. Accurate prediction matters to many, such as clinicians, patients and pharmaceutical companies. For AI research, this challenge is also compelling because it tests whether LLM-based systems can make useful predictions in a domain with immediate and substantial real-world consequences. 

\begin{figure}[tbp]
    \centering
    \includegraphics[width=\linewidth]{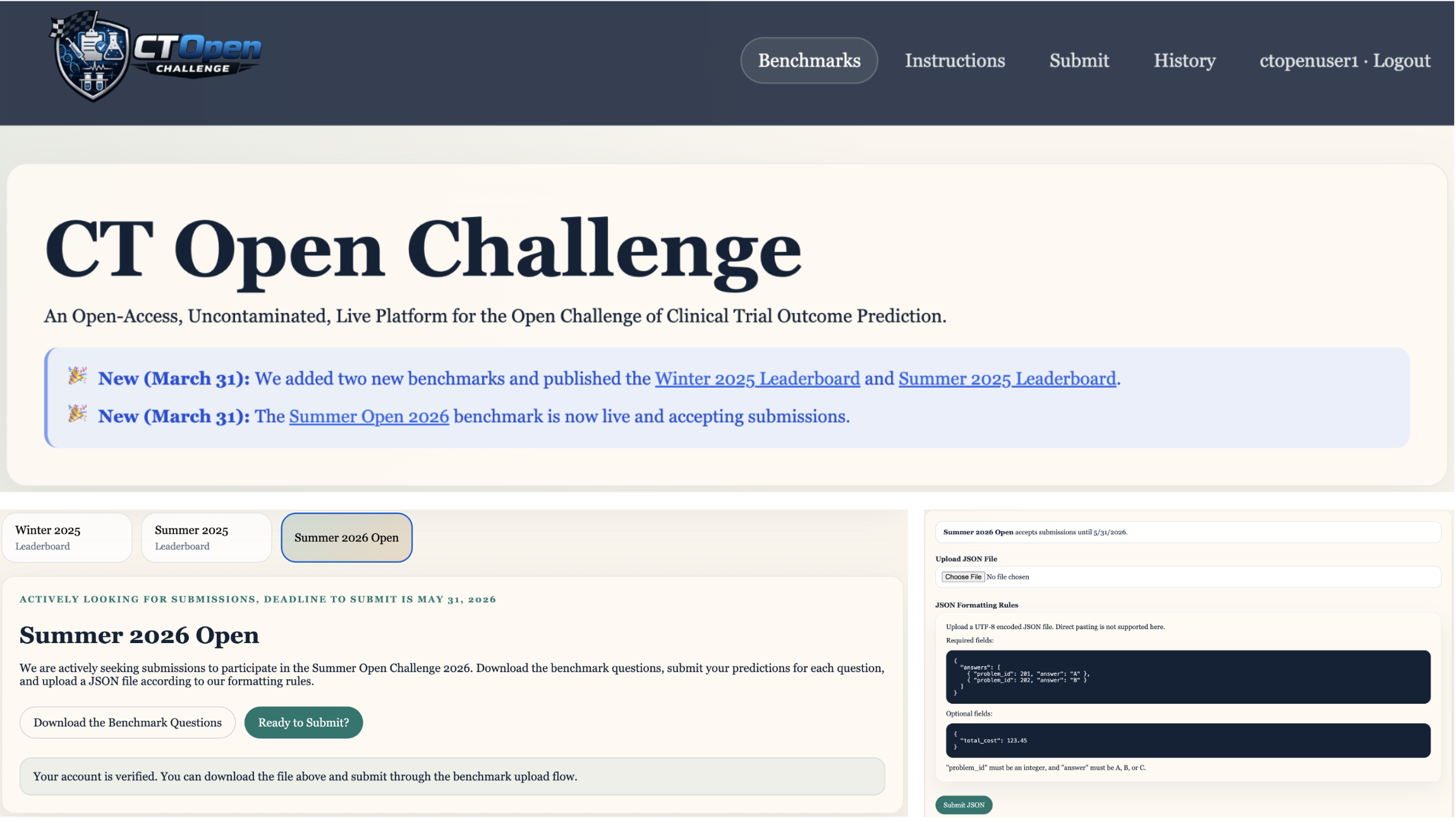}
    \caption{\small CT Open Platform Screenshots. It has released leaderboards for Winter 2025 and Summer 2025 benchmarks. It is now accepting submissions for Summer 2026 Open.}
    \label{fig:ct_open_website}
\end{figure}

To facilitate finding a solution to this challenge, we introduce \textbf{CT Open}, a dynamic platform for clinical trial outcome prediction. See Figure \ref{fig:ct_open_website} for website design. CT Open hosts four challenges each year: Winter Open, Spring Open, Summer Open, and Fall Open. Each challenge corresponds to a three-month window and a fixed set of clinical trial prediction questions. Participants entering a given challenge must submit their predictions before that challenge's window begins. After the window ends, CT Open evaluates all submissions and releases a leaderboard the following week. Specifically, from the pool of clinical trials associated with that challenge’s question set, CT Open identifies the subset of trials that had no public results before the start of the challenge window but acquired usable public results during it. CT Open ranks submissions using only questions from this subset and their derived answers from public results, which are then released as a time-stamped benchmark.

CT Open is distinguished by three properties. First, it is \textbf{dynamic}. Unlike static benchmarks, CT Open will release new time-stamped benchmarks every three-month via its automated, expert-validated pipelines. This helps to evaluate newly released models, while reducing concerns that performance is driven by memorization or benchmark leakage. Second, it is \textbf{open-access} and \textbf{open-method}. Anyone can submit predictions, and participants are not restricted to a predefined evaluation protocol. All methods are permitted. We view this openness as essential: clinical trial prediction is a genuinely difficult open problem, and it is unlikely that a single fixed methodology will be sufficient. Third, CT Open is designed to be \textbf{contamination-resistant}. Our novel and robust decontamination pipeline identifies trials with no publicly available results prior to each challenge's window, including preliminary, interim, subgroup, conference, journal article, news, blog, financial-disclosure evidence, etc. Since submissions are made before each challenge's window, strong performance on CT Open for each challenge is much more plausibly attributable to real predictive ability rather than recall of previously published (partial) results.

Furthermore, CT Open includes a training set of questions and two time-stamped test benchmarks, Winter 2025 and Summer 2025, with cutoff dates of 2025-01-31 and 2025-08-31, respectively. By construction, no clinical trial included in Winter 2025 or Summer 2025 had any publicly available results before its corresponding cutoff date. Therefore, these benchmarks are suitable for evaluating LLMs whose knowledge cutoff, or preferably whose release date, predates the corresponding cutoff timestamp. 

In this paper, we present initial evaluations that illustrate the range of methods CT Open can support. We compare prompt-based LLMs, retrieval-augmented LLMs, agentic LLMs, traditional machine learning methods and neural networks, and find that simple ML and NN baselines remain competitive with, and sometimes outperform, LLM-based approaches. This highlights an important gap in current LLM capabilities: despite their strong performance on many existing benchmarks, they still struggle on difficult predictive tasks such as clinical trial outcome forecasting. In addition, we show that RAG can provide useful evidence from similar completed trials, although its gains are limited and inconsistent, while agentic search can uncover valuable external information but remains computationally expensive and yields mixed results. Finally, by analyzing certain LLM's inconsistent behavior across Winter 2025 and Summer 2025, we show why dynamic, time-stamped, contamination-resistant benchmarks are necessary for fair evaluation. 

\section{Related Work}

\textbf{Clinical trial outcome prediction} has been studied under a range of formulations, from machine learning and neural network methods over structured drug-development and registry features to more recent multimodal methods that incorporate richer trial context \citep{lo2019machine, siah2021predicting, fu2022hint, wang2023spot, zheng2025lifted, hu2025novellanguagemodelpredicting}. Clinical trial benchmarks \citep{chen2025trialbench, gao2024cto} mostly target trial-level result, whereas CT Open consists with fine-grained outcome measure level questions with respect to specific study arms.

\textbf{Retrieval-augmented generation (RAG)} augments an LLM's parametric knowledge with retrieved information at inference time, improving factual grounding while enabling access to updated external information \citep{lewis2021rag}. Subsequent work has expanded this paradigm through iterative, self-reflective, and domain-specialized retrieval strategies \citep{asai2023selfrag, jeong2024selfbiorag, xiong2024imedrag}. This is directly relevant to our setting, where access to trial-relevant evidence such as registry entries, prior publications, and drug- or disease-related context may materially affect outcome prediction.

\textbf{Agentic Systems with Tools and Web Access.} A related line of work studies agentic LLM systems that interleave reasoning with external actions such as web browsing, tool use, API calls, and code execution \citep{nakano2022webgpt, yao2023react, schick2023toolformer, tang-etal-2024-medagents, ClinicalAgent}. Interestingly, most benchmarks, static or dynamic, do not allow agentic system with unrestricted web-search because they can find these published benchmarks and download their answers, while CT Open's unique design makes it suitable to evaluate such systems.

\textbf{LLM Auto-Research and AI Scientist Systems.} Recent work has pushed LLM-based systems beyond single-query question answering toward more autonomous, multi-step research workflows, including literature synthesis, experimentation, and report writing \citep{schmidgall2025agentlaboratoryusingllm, wang2024autosurveylargelanguagemodels, lu2024aiscientistfullyautomated, yamada2025aiscientistv2workshoplevelautomated, gottweis2025aicoscientist}. Our benchmark connects naturally to this line of work by providing an evaluation setting for systems that can run experiments to produce more informed scientific judgments.

\textbf{Dynamic Benchmark.} Most LLM benchmarks are static: they are released once and then reused across successive model generations. Even difficult benchmarks such as GPQA and OlympiadBench become increasingly vulnerable to contamination as newer models are trained on larger web-scale corpora \citep{rein2023gqpa, he2024olympiadbench, balloccu2024leak, deng2024datacontamination}. This has motivated dynamic evaluation, from human-and-model-in-the-loop data collection in Dynabench \citep{kiela2021dynabench} to temporally refreshed benchmarks such as LiveBench \citep{white2025livebench}, LiveCodeBench \citep{jain2024livecodebench}, ForecastBench \citep{karger2025forecastbench}, MathArena \cite{balunovic2026matharena}, and LiveMedBench \citep{yan2026livemedbench}. Our benchmark follows this temporally refreshed perspective for clinical trial outcome prediction by updating quarterly with newly available outcome evidence from the internet, enabling continual evaluation on previously unseen trial results.

\section{Definition of Key Terms}

CT Open focuses on randomized controlled clinical trials that evaluate drug interventions and are registered on \url{clinicaltrials.gov}. We highlight a few important definitions below.

\begin{definition}[NCT ID]
The NCT ID is a trial's identifier on \url{clinicaltrials.gov}.
\end{definition}

\begin{definition}[Intervention/Treatment]
    For CT Open's clinical trials, the intervention (i.e. treatment) is a drug, or a combination of drugs and medical procedures.
\end{definition}

\begin{definition}[Study, Treatment and Comparator Arms]
A study arm is a group of participants assigned to a intervention or control. A treatment arm is a study arm receiving the intervention, while a comparator arm is control, such as placebo.
\end{definition}

\begin{definition}[Outcome Measure and Endpoint]
An outcome measure, or endpoint, is a pre-specified quantity used to evaluate an intervention in a clinical trial. It defines what will be measured, and is part of the trial design rather than the trial result. For example, 5-year overall survival rate is an outcome measure.
\end{definition}

\begin{definition}[Result and Outcome]
A result, or outcome, is the observed value of an outcome measure after the trial is conducted. It describes what actually happened in the trial. For example, 85\% of patients survived for at least 5 years after treatment is a result, whereas 5-year overall survival rate is the corresponding outcome measure.
\end{definition}

\begin{definition}[Challenge, Window, Time-Stamp, Cutoff Date]
A challenge's window is the period of time where we wait for results of clinical trials get published online. A time-stamped benchmark is released the first week after a challenge's window ends. The time-stamp (i.e. the cutoff date) is one day before the challenge's window begins. CT Open ensures no trial in this time-stamped benchmark has any form of results before the cutoff date.
\end{definition}

\section{The Design of CT Open}
CT Open's long-term goal is to facilitate building AI systems that can truly predict clinical trial outcomes, especially at the earliest stage, when a trial is still being planned or has only recently begun. To this end, CT Open uses a novel, robust, carefully-designed \textbf{decontamination pipeline} to ensure that any system evaluated by CT Open cannot rely on partial, interim, or already published trial results, which often appear online in incomplete forms and in scattered places, long before final results are officially posted. 

\subsection{CT Open's Questions}
\label{ssec:ctopen_questions}
CT Open's question is defined at the level of a specific outcome measure of the trial and is represented as a tuple of a clinical trial, a particular outcome measure, and the associated study arms (when applicable). CT Open includes three classes of questions. See the Appendix \ref{sec:example_questions} for example questions and additional details for each class.
\begin{itemize}
    \item The \textbf{Superiority} class consists of binary questions on whether the treatment arm shows a statistically significant improvement over the comparator arm. 
    \item The \textbf{Comparative Effect} class consists of three-way questions on whether the treatment arm is statistically significantly better than, worse than, or not different from the comparator arm. 
    \item The \textbf{Endpoint} class consists of two closely-related binary subtypes of questions: one asks whether the trial meets the endpoint, and the other asks whether at least one arm in the trial meets the endpoint.
\end{itemize} Throughout this paper, meeting an endpoint refers to achieving a statistically significant positive result on an outcome measure. Unless otherwise noted, statistical significance is defined at the 95\% confidence level.

Because clinical trials specify their treatment arms, comparator arms, and outcome measures (i.e. endpoints) even before they start, CT Open can automatically generate questions from the trial information without access to the trial results. When the results are later reported, they can be used to answer those questions via CT Open's \textbf{answer generation pipeline}.

\subsection{The Components of CT Open}
CT Open has two main components: a static component and a dynamic component.

\begin{table}[bpt]
\centering
\footnotesize
\setlength{\tabcolsep}{4pt}
\begin{threeparttable}
\begin{tabular}{lccc}
\toprule
 & \textbf{Winter 2025} & \textbf{Summer 2025} & \textbf{Train} \\
\midrule
\rowcolor{black!4}
Unique Trials & 314 & 240 & 7{,}292 \\
Total Questions & 605 & 857 & 15{,}444 \\
\rowcolor{black!4}
Superiority / Comparative / Endpoint (\%) & 78\% / 9\% / 13\% & 77\% / 6\% / 17\% & 51\% / 31\% / 18\% \\
Trials started in or after 2020 (\%) & 89.1 & 88.4 & 8.8 \\
\rowcolor{black!4}
Median Enrollment & 234 & 212 & 275 \\
Unique Countries & 82 & 81 & 131 \\
\bottomrule
\end{tabular}
\end{threeparttable}
\caption{\small Overall Statistics of the three datasets of CT Open. All three trial sets are disjoint.}
\label{tab:dataset_overview_concise}
\end{table}

As shown in Table \ref{tab:dataset_overview_concise}, the static component consists of a training set of 7292 unique clinical trials with officially reported results, together with approximately 15,444 questions derived from those trials. The static component also includes two time-stamped test benchmarks, Winter 2025 (605 questions) and Summer 2025 (857 questions), with respective time-stamps at 2025-01-31 and 2025-08-31. They encompass a diverse range of biological and medical topics, including oncology, cardiovascular diseases, autoimmune disorders and neurology. For details, refer to Table \ref{tab:dataset_overview} and Table \ref{tab:disease_dist} in Appendix \ref{subsec:dataset_stats}.

The dynamic component consists of a series of time-stamped benchmarks released every three months. The first benchmark is scheduled to be constructed via CT Open's \textbf{decontamination pipeline} and \textbf{answer generation pipeline}, and released on 2026-09-07 and will use a cutoff date of 2026-05-31. Future benchmarks will be constructed and released on the same quarterly schedule. Note the Winter 2025 benchmark has a different cutoff date.

\subsection{The Schedule of CT Open}
\begin{table}[hpbt]
\centering
\setlength{\tabcolsep}{8pt}
\small
\begin{tabular}{llll}
\toprule
\textbf{Challenge} & \textbf{Window} & \textbf{Submission Deadline} & \textbf{Leaderboard Release} \\
\midrule
Winter Open & December--March & November 30 & First week of March \\
Spring Open & March--June & Last day of February & First week of June \\
Summer Open & June--September & May 31 & First week of September \\
Fall Open & September--December & August 31 & First week of December \\
\bottomrule
\end{tabular}
\caption{\small CT Open hosts four challenges every year. Submission begins when the question pool for each challenge is released, usually 2-3 months before deadline.}
\label{tab:ctopen_cycles}
\end{table}

CT Open hosts four challenges each year: Winter Open, Spring Open, Summer Open, and Fall Open (Table~\ref{tab:ctopen_cycles}). The CT Open website is scheduled to go live in the first week of April 2026, so the first active challenge will be Summer Open 2026. Participants may submit predictions until May 31, 2026, and the leaderboard will be released in the first week of September 2026. Submissions received after that deadline will be evaluated in the Fall Open.

For each challenge, CT Open releases a downloadable question set derived from trials that are likely to report results within that challenge’s window. At the end of the window, CT Open runs its \textbf{answer generation pipeline} and \textbf{decontamination pipeline} to identify the subset of trials with usable results posted during that window and no public results posted before the window begins. Trials that fail this condition are considered unusable or contaminated, and are removed. The remaining trials, and their questions and answers form the time-stamped benchmark for that challenge, which is then used to evaluate submissions and produce the leaderboard.

Importantly, CT Open only evaluates submissions made before a challenge window begins, and it excludes any clinical trial for which results are publicly available before that window. As a result, systems evaluated on CT Open cannot benefit from prior access to the outcomes of the trials used for evaluation. Therefore, any substantial improvement in performance is more likely to reflect genuine predictive ability rather than retrieval or recall of existing results. 

\subsection{The Decontamination Pipeline}
To construct a time-stamped benchmark for each challenge, CT Open applies a novel and robust decontamination pipeline to identify suitable clinical trials from the pool of trials associated with that challenge’s question set. This pipeline combines multiple search mechanisms, including LLM-based web search and a customized web and database search system, to detect publicly available trial results and exclude contaminated trials.

We define the following notation: $$\textsc{Search}(query, [start,end], mode) \longrightarrow \{doc_1, doc_2...\}$$ as a search procedure that uses \textit{query} to retrieve all documents about a clinical trial, subject to the publication-date constraint $[\textit{start}, \textit{end}]$. Each returned document must be verifiably about the queried trial and must report trial results. A document can take any form: journal article, press release, financial statement, blog posts, etc. The \textit{query} may be a keyword (e.g. trial's NCT ID), a sentence-length query or a multi-paragraph description. If \textit{start} date is $-\infty$, no lower bound is imposed on publication date; if \textit{end} date is $+\infty$, results may be as recent as the day of search.

The parameter \textit{mode} specifies the retrieval system. If \textit{mode} is an LLM (e.g., GPT-5.2 or Gemini 3), search is performed through the corresponding model's web-search capability. If \textit{mode} is Brave, we use a customized pipeline in which Brave Search API returns candidate URLs, webpages are fetched with Python Requests or ZenRows, and downloadable files are converted to text with DeepSeek OCR \citep{wei2025deepseekocr}. Scraped webpages are then processed with a customized text-extraction system derived from trafilatura \citep{barbaresi-2021-trafilatura}, designed for long pages and for isolating the main article content while removing HTML, sidebars, and other articles' text. If \textit{mode} = Database, we use the trial's NCT ID to search PubMed, PMC, BioRxiv, and MedRxiv for matching articles. If \textit{mode} is Brave or Database, all retrieved texts, whether from webpages, downloadable files, or databases, first undergo our customized pipeline for published date extraction to verify its within [$start$, $end$], and then undergo two rounds of GPT-5 verification: first, to confirm that the document refers to the same clinical trial; and second, to confirm that it reports some form of results for that trial. See Figure \ref{fig:clinical_trial_identification_prompt} and Figure \ref{fig:result_verification_prompt} for prompts.

\begin{algorithm}
\small
\setstretch{1.1}
\caption{\small Filtering trials with pre-cutoff results or no post-cutoff results}
\label{alg:decontamination}
\KwIn{Initial trial pool $\mathcal{P}$, cutoff date $c$}
\KwOut{Filtered trial set $\mathcal{F}$}

$\mathcal{F} \leftarrow \emptyset$\;

\ForEach{trial $\in \mathcal{P}$}{
    $\mathrm{info} \leftarrow \{\text{NCT ID, aliases, title, description}\}$\;

    $U_{\text{before}} \leftarrow
    \textsc{SEARCH}(\mathrm{info}, [-\infty, c), \text{GPT-5.2})
    \cup
    \textsc{SEARCH}(\mathrm{info}, [-\infty, c), \text{Gemini 3})$\;
    \lIf{$U_{\text{before}} \neq \emptyset$}{\textbf{continue}} 

    $U_{\text{after}} \leftarrow
    \textsc{SEARCH}(\mathrm{info}, [c, +\infty), \text{GPT-5.2})
    \cup
    \textsc{SEARCH}(\mathrm{info}, [c, +\infty), \text{Gemini 3})$\;
    \lIf{$U_{\text{after}} = \emptyset$}{\textbf{continue}}

    Generate three rewritten queries $Q_1, Q_2, Q_3$ from $\mathrm{info}$ using GPT5\;

    $U_{\text{brave}} \leftarrow \bigcup_{i \in \{1,2,3\}} \textsc{SEARCH}(Q_i, [-\infty, c), \text{Brave})$\;
    $U_{\text{db}} \leftarrow \textsc{SEARCH}(\text{NCT ID}, [-\infty, c), \text{Database})$\;
    \lIf{$U_{\text{brave}} \neq \emptyset$ \textbf{or} $U_{\text{db}} \neq \emptyset$}{\textbf{continue}}

    add $t$ to $\mathcal{F}$\;
}

\Return{$\mathcal{F}$}
\end{algorithm}

Algorithm \ref{alg:decontamination} summarizes our decontamination pipeline. To assess its robustness, we performed an additional validation using Embase, a proprietary Elsevier database that includes biomedical literature and conference abstracts not necessarily accessible through public search. Because our pipeline relies entirely on publicly available information, Embase provides a useful external check for residual contamination. We applied this validation to all clinical trials that passed our pipeline for both the Winter 2025 and Summer 2025 benchmarks and found no trial with reported results in Embase prior to the corresponding cutoff date. This shows our pipeline is robust even when it can only access public information.

We also randomly sampled 50 NCT IDs for manual review. For each sampled trial, one researcher conducted extensive search using Google Search with manually designed queries, coupled with GPT-5 web search, and GPT-5 Deep Research as supplements, examining more than 200 webpages per trial. For each trial, this process could take 2 to 4 hours. Our annotators confirmed that 49 of the 50 sampled NCT IDs had no reported results before their respective cutoff dates. Only one trial was classified as ambiguous. Our annotator identified a 2024 LinkedIn post by the trial’s principal investigator that included a blurry photograph of a conference poster appearing to show partial trial results. This single image is the only evidence available prior to the cutoff date, and, to the best of our knowledge, current LLMs are pretrained on textual data. To check the impact of this image, we showed that o3-mini, GPT-5, Opus 4.5 Qwen-2.5-14B and DeepSeek-Distilled-Qwen-2.5-14B have no recall of this image's information. GPT-5 and Opus 4.5 could not use web-search to find it. We asked each model to predict the trial's outcome and inspected its reasoning traces; none referenced or alluded to the poster.

Nevertheless, conservatively, the decontamination pipeline achieves an accuracy of at least $98\%$. As a precaution, we exclude this trial and its associated questions from the test benchmarks.

\subsection{Answer Generation Pipeline}

Each remaining trial is associated with a set of result documents published within the challenge window. We design a multi-stage GPT-5.4-based pipeline to determine whether the reported results are statistically rigorous; can be clearly linked to specific outcome measures and study arms; and are sufficient to support answering the questions associated with that trial. If the pipeline determines that the available results cannot support a question with sufficient certainty, that question is removed from the benchmark. See Figure \ref{fig:answer_verification_endpoint_superiority}  and Figure \ref{fig:comparative_effect_answer_verification}  for prompts. 

To validate this pipeline, we randomly sampled 100 questions, answers and reported results. We then asked a physician-scientist experienced in clinical trial design and analysis to annotate these 100 cases and assess whether the proposed answers were supported by the reported results. The expert's annotation shows an 99\% accuracy.

\begin{figure}[t]
    \centering
    \includegraphics[width=1.0\linewidth]{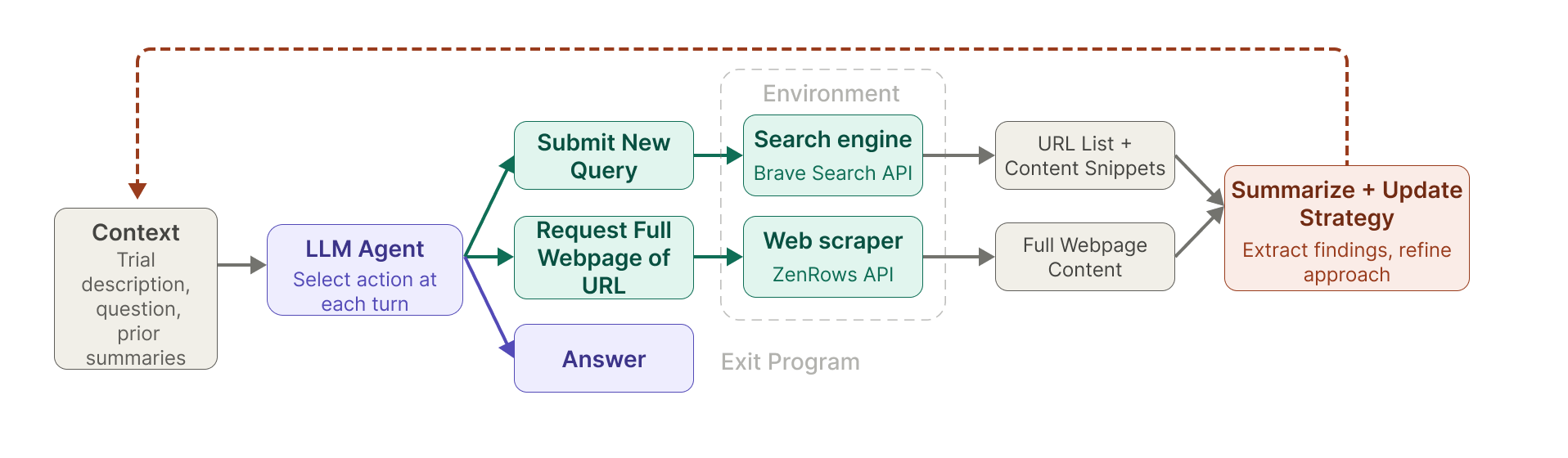}
    \caption{\small Agentic LLM Workflow}
    \label{fig:agentic_llm}
\end{figure}

\begin{table*}[t]
\centering
\small
\setlength{\tabcolsep}{10pt}
\renewcommand{\arraystretch}{1.10}
\begin{threeparttable}
\begin{tabular}{lcccccc}
\toprule
\multirow{2}{*}{\textbf{Model}} &
\multicolumn{2}{c}{\textbf{Endpoint}} &
\multicolumn{2}{c}{\textbf{Superiority}} &
\multicolumn{2}{c}{\textbf{Comparative Effect}} \\
\cmidrule(lr){2-3}
\cmidrule(lr){4-5}
\cmidrule(lr){6-7}
& \textbf{Macro-F1} & \textbf{Acc}
& \textbf{Macro-F1} & \textbf{Acc}
& \textbf{Macro-F1} & \textbf{Acc} \\
\midrule

\rowcolor{black!4}
\multicolumn{7}{l}{\textbf{Winter 2025}} \\
\midrule

\multicolumn{7}{l}{\textit{Traditional Baselines}} \\
Random Forest                & 66.60 & 71.54 & 55.39 & 63.76 & \best{80.70} & \best{79.12} \\
Feed-Forward NN                & 67.74 & 72.29 & 54.23 & 57.91 & 58.87 & \second{74.60} \\
\midrule

\multicolumn{7}{l}{\textit{Large Language Models}} \\
o3-mini                        & 68.40 & 70.23 & \second{68.09} & 68.33 & 51.91 & 51.53 \\
\hspace{1em} + RAG             & \best{76.88}\sigup & 73.48\nomup & \best{69.21}\nomup & 69.18\nomup & 56.81\nomup & 57.14\nomup \\
\hspace{1em} + Agent           & 64.83\nomdown & 66.67\nomdown & 67.46\nomdown & 68.06\nomdown & 54.74\nomup & 55.72\nomup \\
GPT-5                          & 65.29 & 77.63 & 66.17 & \second{71.20} & 54.25 & 54.70 \\
\hspace{1em} + RAG             & 67.98\nomup & 78.03\nomup & 67.54\nomup & \best{71.86}\nomup & 56.96\nomup & 57.49\nomup \\
Claude Opus 4.5$^{\dagger}$               & 70.17 & \second{80.23} & 62.31 & 68.91 & 55.52 & 54.83 \\
\hspace{1em} + RAG             & \second{70.36}\nomup & \best{81.85}\nomup & 63.43\nomup & 69.62\nomup & 56.00\nomup & 55.19\nomup \\
\midrule
\multicolumn{7}{l}{\textit{Open-Source Large Language Models}} \\
Qwen2.5-14B                    & 53.29 & 53.68 & 60.04 & 60.21 & 58.00 & 62.94 \\
\hspace{1em} + RAG             & 68.77\nomup & 66.27\nomup & 62.67\nomup & 62.34\nomup & \second{60.32}\nomup & 60.64\nomdown \\
\hspace{1em} + Agent           & 61.41\nomup & 55.63\nomup & 59.74\nomdown & 61.53\nomup & 34.20\nomdown & 33.78\nomdown \\
DeepSeek-R1-14B                & 53.54 & 54.18 & 56.68 & 58.02 & 45.69 & 55.05 \\
\hspace{1em} + RAG             & 67.56\nomup & 64.21\nomup & 61.26\nomup & 61.09\nomup & 32.30\nomdown & 41.31\nomdown \\
\hspace{1em} + Agent           & 66.23\nomup & 64.07\nomup & 61.21\nomup & 61.77\nomup & 51.21\nomup & 56.78\nomup \\
\midrule
\rowcolor{black!4}

\rowcolor{black!4}
\multicolumn{7}{l}{\textbf{Summer 2025}} \\
\midrule

\multicolumn{7}{l}{\textit{Traditional Baselines}} \\
Random Forest                  & \best{65.57} & \second{66.48} & 55.88 & 61.63 & 46.15 & 45.65 \\
Feed-Forward NN                & \second{65.25} & \best{67.78} & 64.95 & 67.58 & 57.56 & \second{85.87} \\
\midrule

\multicolumn{7}{l}{\textit{Large Language Models}} \\
o3-mini                        & 59.83 & 59.46 & 72.48 & 71.93 & 54.94 & 59.78 \\
\hspace{1em} + RAG             & 59.07\nomdown & 58.98\nomdown & \second{73.15}\nomup & 72.94\nomup & 56.75\nomup & 60.87\nomup \\
\hspace{1em} + Agent           & 61.75\nomup & 62.84\nomup & \best{73.33}\nomup & \second{73.96}\nomup & \best{69.81}\sigup & \best{92.75}\sigup \\
GPT-5                          & 51.54 & 58.24 & 70.23 & 73.71 & 64.81 & 64.49 \\
\hspace{1em} + RAG             & 50.13\nomdown & 55.55\nomdown & 70.18\nomdown & 73.78\nomup & \second{68.95}\nomup & 70.05\nomup \\
Claude Opus 4.5                & 53.71 & 57.86 & 68.76 & 73.68 & 48.24 & 49.64 \\
\hspace{1em} + RAG             & 58.69\sigup & 62.72\sigup & 69.59\nomup & \best{74.10}\nomup & 48.42\nomup & 50.00\nomup \\
\midrule
\multicolumn{7}{l}{\textit{Open-Source Large Language Models}} \\
Qwen2.5-14B                    & 53.42 & 53.61 & 58.84 & 59.43 & 44.74 & 44.69 \\
\hspace{1em} + RAG             & 58.91\nomup & 58.55\nomup & 61.00\nomup & 61.02\nomup & 45.35\nomup & 44.20\nomdown \\
\hspace{1em} + Agent           & 60.60\nomup & 59.82\nomup & 55.13\nomdown & 57.85\nomdown & 32.14\nomdown & 23.68\nomdown \\
DeepSeek-R1-14B                & 52.04 & 52.25 & 57.95 & 59.65 & 41.39 & 51.21 \\
\hspace{1em} + RAG             & 58.32\nomup & 57.83\nomup & 61.48\nomup & 61.79\nomup & 32.92\nomdown & 47.71\nomdown \\
\hspace{1em} + Agent           & 57.31\nomup & 56.90\nomup & 57.59\nomdown & 58.33\nomdown & 29.63\nomdown & 21.05\nomdown \\
\bottomrule
\end{tabular}
\end{threeparttable}
\begin{tablenotes}[flushleft]
\footnotesize
\item[$\dagger$] $^{\dagger}$ Opus 4.5 knowledge cutoff date postdates Winter 2025's timestamp, possibly contaminated.
\end{tablenotes}
\caption{\small
Acc is balanced accuracy. Compared to base model, \textnomup and \textnomdown indicate nominal increases and decreases, while \textsigup denotes statistically significant improvements (95\% CI via hierarchical bootstrap sampling). Best and second-best results within each dataset and column are marked in \best{bold} and \second{underline}, respectively. All LLM results are averaged over three runs.}
\label{tab:main_eval_results_concise}
\end{table*}

\section{Evaluation}
We evaluate o3-mini \citep{openai2025o3mini}, GPT-5 \citep{singh2025openaigpt5card}, Claude Opus 4.5 \citep{anthropic2025opus45} using the following approaches on Winter 2025 and Summer 2025. All prompts, agent protocols, and hyperparameters for the traditional machine learning and neural network methods are developed on the validation set, which is 10\% of the training set.

\textbf{Prompt-based LLM approach:} We provide the model with the trial title, trial description, drug description when available, patient eligibility criteria, the question and its outcome measure, and the relevant study arms when available. The model is then asked to predict the most likely answer choice. We found that adding numerical and categorical variables reduced performance, so they were excluded. See Appendix \ref{subsec:evaluation_prompts} Figure \ref{fig:baseline_evaluation_prompt} for prompt.

\textbf{Machine Learning and Neural Network methods:} We train on 90\% of the training set and reserve the remaining 10\% for hyperparameter tuning. We include all available categorical and numerical trial features, such as phase and enrollment. See Appendix \ref{subsec:baseline_implementation_details} for details.

\textbf{Agentic LLM:} We design an iterative search-and-browse workflow. At each step, the agent chooses either to issue a new query to a Search engine or to inspect the full content of a previously retrieved URL. Query results are filtered to include only documents whose publication dates are verified to precede the benchmark cutoff, and the agent receives the corresponding URLs together with their content snippets. After each step, the agent summarizes the newly observed information and may revise its search strategy; only these summaries, rather than the full retrieved content, are retained in context. Agent can interact with the environment for up to 25 rounds before producing answer. See Figure \ref{fig:agentic_llm} for agentic workflow and see Appendix \ref{subsec:evaluation_prompts} for agent protocol in Figure \ref{fig:agent_evaluation_protocol}. 

\textbf{LLM with RAG:} In preliminary experiments, retrieving preclinical evidence, such as animal and observational studies, consistently degraded baseline LLM performance on the validation set. We found that retrieval is more useful when it focuses on previously completed clinical trials with publicly available results that closely resemble the target trial in drug mechanism, design, outcome measure, and patient population. Because these prior trials are not exact matches, their outcomes cannot be used by simple aggregation; rather, the model must reason about how small differences in trial design may affect the final result. Motivated by this observation, we developed a customized retrieval pipeline to get a list of similar completed trials, averaging about 15 per target trial, each paired with its most relevant endpoint. We then provide the LLM with these similar trials and their reported results from \url{clinicaltrials.gov}, and ask it to reason over the differences between the retrieved trials and the target trial before predicting the final answer. See Figure \ref{fig:RAG_similar_evaluation_prompt}, \ref{fig:RAG_metoo_evaluation_prompt}, \ref{fig:RAG_no_match_evaluation_prompt} for prompts.

\section{Analyses and Future Directions}

\textbf{Traditional ML and NN methods remain competitive with LLM-based approaches:} This suggests that these models can effectively learn correlations between categorical, numerical features and the target labels. By contrast, off-the-shelf LLMs appear less effective at using such variables, consistent with our initial experiments. These findings point to a promising direction for future work: improving LLMs’ ability to learn and exploit correlational structure in tabular and other structured data, or enabling LLM-driven systems to train more effective classical predictors, such as random forests and lightweight neural networks, in an auto-research or AI-scientist-style setting. More broadly, the results highlight a persistent challenge for LLM research: although modern LLMs perform strongly on many tasks that humans can already do, they still struggle on open predictive problems that remain difficult for humans themselves. For brevity, Table \ref{tab:main_eval_results_concise} reports only the strongest traditional baseline; full baseline results are provided in Table \ref{tab:main_eval_results_full} in the Appendix \ref{sec:traditional_baseline_details}. A grouped permutation-importance analysis of which feature groups these baselines rely on is provided in Appendix~\ref{subsec:feature_importance}.

\textbf{RAG yields modest improvements over the baseline in certain cases}: We manually examined cases in which the LLM with RAG produced an incorrect answer while the baseline LLM answered correctly. In most such cases, the retrieved similar trials were not themselves irrelevant; rather, the model appeared unable to properly account for subtle but consequential differences between the retrieved trials and the target trial. Nevertheless, for question classes where RAG did improve performance, for example, comparing o3-mini with o3-mini + RAG on Winter 2025, and Opus 4.5 with Opus 4.5 + RAG on Summer 2025, the reported outcomes of similar prior trials appear to provide a useful starting point for reasoning about the target trial and forming a prediction.

\textbf{The Agent setting produces mixed results:} We found the agent setting to be both computationally expensive and double-edged. Running the current agent configuration with o3-mini costs more than \$1000 in tokens, so we were only able to evaluate it with o3-mini. On Winter 2025, o3-mini with the agent setting nominally underperformed the o3-mini baseline. On Summer 2025, however, o3-mini with the agent setting outperformed the baseline, particularly on Comparative Effect questions. Manual inspection suggests that the agent can sometimes discover unexpected but highly informative sources that substantially alter its search strategy. For example, in one case it encountered a \url{withpower.com} result, which prompted the agent to include \texttt{withpower} in a subsequent query, allowing Brave Search to target the site directly. Because WithPower contains relevant information and analyses on clinical trials, the agent was able to extract additional useful evidence from related trials hosted there. In another case, the agent discovered \url{https://copingwithmsa.com/}, a patient-focused website containing  firsthand accounts related to clinical trials. By refining its queries to search that site more deeply, the agent located a relevant patient blog describing experiences with a similar trial, which provided useful evidence for prediction.

\textbf{Open-weight models benefit from retrieval but not from the agent setting:} Qwen2.5-14B and DeepSeek-R1-14B score 52.04-53.54 macro-F1 on Endpoint in both benchmarks, close to random assignment on a binary task. RAG improves both models on the two binary tasks, with the largest gains on Winter 2025 Endpoint (+15.48 and +14.02 macro-F1), but its effect on three-way Comparative Effect is mixed. The agent setting improves both models on Endpoint only and lowers Qwen2.5-14B's macro-F1 on the other two tasks in both benchmarks. Error analysis with our biomedical expert identified two failure modes. First, the models retrieve relevant prior trials without adjusting for design differences: on NCT06360458 (MIRACLE), Qwen2.5-14B retrieved the comparable MARVEL trial but did not account for MIRACLE enrolling the subgroup MARVEL explicitly excluded. Second, despite explicit instruction otherwise, both models spent much of their query budget searching for unpublished outcomes that our decontamination pipeline ensures were unavailable before the cutoff date, providing further evidence that the benchmark is not solvable by retrieval.

\textbf{Pretrained knowledge may affect performance on contaminated benchmarks:} We intentionally evaluated Opus 4.5 on Winter 2025, whose cutoff date is January 31, 2025, even though Opus 4.5 has a knowledge cutoff of August 31, 2025. On Winter 2025, Opus 4.5 outperformed o3-mini and GPT-5 in five settings: Endpoint macro-F1 and accuracy, Comparative Effect macro-F1 and accuracy, and Superiority accuracy. By contrast, on Summer 2025, whose cutoff date is August 31, 2025 and is therefore technically uncontaminated with respect to Opus 4.5’s knowledge cutoff, Opus 4.5 underperformed o3-mini and GPT-5 in four settings: Endpoint accuracy, Superiority macro-F1, and Comparative Effect macro-F1 and accuracy. These results illustrate the importance of dynamic, uncontaminated benchmarks, which enable recently released models such as Opus 4.5 to be evaluated more fairly and reward genuine predictive ability rather than recall from pretraining data.

\section{Conclusion}
In this paper, we present CT Open and its automated, expert-validated benchmark construction algorithms. The design of CT Open will ensure that it is contamination-resistant and that it would properly differentiate systems that display genuine predictive abilities over systems that are over-fitted or contaminated. CT Open will serve as a central hub that brings together the community interested in clinical trial outcome prediction or advancing the capabilities of LLMs. Our analyses have revealed critical gaps of current RAG and agentic systems for performing predictive tasks, which highlights many opportunities to further improve these systems, and even potentially conceive new methodology. 

\section*{Ethics Statement}
CT Open does not and will never publish patient's personal information or display them on the platform. All questions, answers, and evaluations use publicly available information that is aggregate statistics, and not individual's information. All clinical trials at CT Open are approved and standard trials registered at \url{clinicaltrials.gov}.

\bibliography{colm2026_conference}
\bibliographystyle{colm2026_conference}
\newpage
\clearpage

\newpage
\appendix

\textbf{Appendix: Table of Contents}
    \begin{enumerate}[label=\Alph*.]
        \item Dataset Statistics and Full Evaluation Results \dotfill \pageref{subsec:dataset_stats}
        \item Traditional Baseline Implementation Details \dotfill \pageref{sec:traditional_baseline_details}
            \begin{enumerate}[label=\arabic*.]
                \item Full data feature set for tabular machine learning models \dotfill \pageref{subsec:full_data_feature_set}
                \item Baseline Implementation Details and Hyperparameters \dotfill \pageref{subsec:baseline_implementation_details}
                \item Feature Group Importance Analysis \dotfill \pageref{subsec:feature_importance}
            \end{enumerate}
        \item Pipelines \dotfill \pageref{subsec:pipelines}
            \begin{enumerate}[label=\arabic*.]
                \item Decontamination Pipeline Implementation Detail \dotfill \pageref{subsec:decontamination_detail}
                \item Answer Generation Pipeline Implementation Detail \dotfill \pageref{subsec:answer_generation_detail}
            \end{enumerate}
        \item Dataset Generation Prompts \dotfill \pageref{subsec:dataset_generation_prompts}
        \item Evaluation Prompts \dotfill \pageref{subsec:evaluation_prompts}
        \item Question Classes, Label Distribution, and Example Questions \dotfill \pageref{sec:example_questions}
    \end{enumerate}
\clearpage

\clearpage
\appendix

\section{Dataset Statistics and Full Evaluation Results}
\label{subsec:dataset_stats}
Tables~\ref{tab:dataset_overview} and~\ref{tab:disease_dist} summarize 
the scale, temporal distribution, and disease-area composition of each  data split. Table~\ref{tab:main_eval_results_full} presents the full evaluation results across all baselines and models. Table~\ref{tab:label_dist} in Appendix~\ref{sec:example_questions} reports the label distribution within each question class.

\begin{table*}[ht]
\centering
\footnotesize
\setlength{\tabcolsep}{10pt}
\begin{tabular}{@{}l@{\hskip 12pt}rrr@{}}
\toprule
& \textbf{Winter 2025} & \textbf{Summer 2025} & \textbf{Train} \\
\midrule
\rowcolor{gray!6}
\multicolumn{4}{@{}l}{\textbf{Scale}} \\
\quad Unique trials                    & 314           & 240           & 7{,}292       \\
\quad (Trial, endpoint) pairs          & 561           & 781           & 10{,}842      \\
\quad Total questions                  & 605       &  857      & 15{,}444     \\
\quad Superiority / Comparative / Endpoint (\%) & 78 / 9 / 13   & 77 / 6 / 17   & 51 / 31 / 18  \\
\quad Primary / Secondary / Other (\%) & 54 / 44 / 2   & 34 / 65 / 1   & 32 / 66 / 2   \\
\midrule
\rowcolor{gray!6}
\multicolumn{4}{@{}l}{\textbf{Temporal distribution} {\normalfont\textit{(study start year)}}} \\
\quad Year range                       & 2011--2025    & 2009--2025    & 1989--2023    \\
\quad $<$\,2020                        & 35\phn (11.1\%)   & 25\phn (10.4\%)   & 6{,}649 (91.2\%) \\
\quad 2020--2023                       & 229 (72.9\%)  & 131 (54.6\%)  & 625\phn\phn (8.6\%)   \\
\quad 2024--2025                       & 51\phn (16.2\%)   & 81\phn (33.8\%)   & 0\phn\phn\phn\phn (0.0\%)   \\
\midrule
\rowcolor{gray!6}
\multicolumn{4}{@{}l}{\textbf{Enrollment}} \\
\quad Median / Mean                    & 234 / 922     & 212 / 382     & 275 / 660     \\
\quad Range                            & 50--149{,}201 & 50--6{,}300   & 50--144{,}000 \\
\midrule
\rowcolor{gray!6}
\multicolumn{4}{@{}l}{\textbf{Trial phase}} \\
\quad Phase 3                          & 81 (57.6\%)  & 130 (54.2\%)  & 4{,}250 (58.3\%) \\
\quad Phase 2/3                        & 213\phn (7.3\%)    & 12\phn (5.0\%)    & 259\phn\phn (3.6\%)   \\
\quad Phase 2                          & 91\phn (29.0\%)   & 78\phn (32.5\%)   & 2{,}534 (34.7\%) \\
\quad Phase 1/2                        & 3\phn\phn (1.0\%)     & 3\phn\phn (1.3\%)     & 159\phn\phn (2.2\%)   \\
\quad Phase 1 / Early                  & 16\phn (5.1\%)    & 17\phn (7.1\%)    & 90\phn\phn\phn (1.2\%)    \\
\midrule
\rowcolor{gray!6}
\multicolumn{4}{@{}l}{\textbf{Geography}} \\
\quad Unique countries                 & 82            & 81            & 131           \\
\quad Top-3 countries                  & US, ES, CN    & US, CN, ES    & US, CA, DE    \\
\quad Trials w/ US sites               & 181 (57.6\%)  & 115 (47.9\%)  & 5{,}282 (72.4\%) \\
\midrule
\rowcolor{gray!6}
\multicolumn{4}{@{}l}{\textbf{Intervention type}} \\
\quad Drug                             & 690 (84.3\%)  & 519 (80.2\%)  & 17{,}814 (84.6\%) \\
\quad Biological                       & 50\phn (6.1\%)    & 67\phn (10.4\%)   & 1{,}536\phn (7.3\%)  \\
\quad Other                            & 78\phn (9.5\%)    & 61\phn\phn (9.4\%)    & 1{,}738\phn (8.3\%)  \\
\bottomrule
\end{tabular}
\caption{\small Dataset overview and key characteristics. All trials are interventional with no shared NCT IDs across any pair of splits. A trial may map to multiple disease categories.}
\label{tab:dataset_overview}
\end{table*}

\begin{table}[ht]
\centering
\begin{threeparttable}
\footnotesize
\setlength{\tabcolsep}{28pt}
\begin{tabular}{@{}lrrr@{}}
\toprule
\textbf{Disease Area} & \textbf{Winter 2025} & \textbf{Summer 2025} & \textbf{Train} \\
\midrule
Oncology               & 24.5 & 21.3 & 21.1 \\
Cardiovascular         & 13.7 &  8.3 &  5.9 \\
Metabolic / Endocrine  &  8.3 & 14.2 & 10.7 \\
Autoimmune / Inflamm.  &  8.0 &  9.6 &  6.7 \\
Neurology / Psychiatry &  7.3 &  9.2 &  9.6 \\
Infectious Disease     &  6.4 &  6.7 &  8.7 \\
Rare / Genetic         &  5.7 &  2.5 &  4.3 \\
Respiratory            &  4.5 &  4.6 &  7.2 \\
Dermatology            &  4.5 &  3.3 &  2.9 \\
Pain / Anesthesia      &  4.5 &  2.5 &  4.5 \\
Nephrology             &  3.8 &  3.8 &  2.5 \\
Musculoskeletal        &  3.5 &  3.3 &  4.7 \\
\midrule
Other\textsuperscript{\textdagger}  & 15.6 & 19.6 & 20.8 \\
\bottomrule
\end{tabular}
\begin{tablenotes}[flushleft]
\scriptsize
\item[\textdagger] Includes hematology, gastroenterology, ophthalmology, and unmatched conditions.
\end{tablenotes}
\end{threeparttable}
\caption{\small Disease-area distribution (\% of unique trials). Categories below 2\% in all splits are grouped under ``Other.''}
\label{tab:disease_dist}
\end{table}

\begin{table*}[t]
\centering
\small
\setlength{\tabcolsep}{8.6pt}
\renewcommand{\arraystretch}{1.10}
\begin{threeparttable}
\begin{tabular}{lcccccc}
\toprule
\multirow{2}{*}{\textbf{Model}} &
\multicolumn{2}{c}{\textbf{Endpoint}} &
\multicolumn{2}{c}{\textbf{Superiority}} &
\multicolumn{2}{c}{\textbf{Comparative Effect}} \\
\cmidrule(lr){2-3}
\cmidrule(lr){4-5}
\cmidrule(lr){6-7}
& \textbf{Macro-F1} & \textbf{Acc}
& \textbf{Macro-F1} & \textbf{Acc}
& \textbf{Macro-F1} & \textbf{Acc} \\
\midrule

\rowcolor{black!4}
\multicolumn{7}{l}{\textbf{Winter 2025}} \\
\midrule

\multicolumn{7}{l}{\textit{Traditional Baselines}} \\
Random Forest                  & 66.60 & 71.54 & 55.39 & 63.76 & \best{80.70} & \best{79.12} \\
Feed-Forward NN                & 67.74 & 72.29 & 54.23 & 57.91 & 58.87 & \second{74.60} \\
KNN + Random Forest            & 65.49 & 70.78 & 54.87 & 63.55 & \second{75.45} & 72.87 \\
L2 Logistic Regression         & 62.38 & 70.56 & 54.46 & 58.58 & 65.01 & 67.55 \\
HINT                           & 57.28 & 64.72 & 51.87 & 62.33 & 46.08 & 50.00 \\
AdaBoost                       & 67.55 & 70.24 & 54.85 & 58.41 & 50.45 & 64.10 \\
XGBoost                        & 65.52 & 74.89 & 58.29 & 63.47 & 69.47 & 66.62 \\
\midrule

\multicolumn{7}{l}{\textit{Large Language Models}} \\
o3-mini                        & 68.40 & 70.23 & \second{68.09} & 68.33 & 51.91 & 51.53 \\
\hspace{1em} + RAG             & \best{76.88}\sigup & 73.48\nomup & \best{69.21}\nomup & 69.18\nomup & 56.81\nomup & 57.14\nomup \\
\hspace{1em} + Agent           & 64.83\nomdown & 66.67\nomdown & 67.46\nomdown & 68.06\nomdown & 54.74\nomup & 55.72\nomup \\
GPT-5                          & 65.29 & 77.63 & 66.17 & \second{71.20} & 54.25 & 54.70 \\
\hspace{1em} + RAG             & 67.98\nomup & 78.03\nomup & 67.54\nomup & \best{71.86}\nomup & 56.96\nomup & 57.49\nomup \\
Claude Opus 4.5$^{\dagger}$                & 70.17 & \second{80.23} & 62.31 & 68.91 & 55.52 & 54.83 \\
\hspace{1em} + RAG             & \second{70.36}\nomup & \best{81.85}\nomup & 63.43\nomup & 69.62\nomup & 56.00\nomup & 55.19\nomup \\
\midrule
\multicolumn{7}{l}{\textit{Open-Source Large Language Models}} \\
Qwen2.5-14B                    & 53.29 & 53.68 & 60.04 & 60.21 & 58.00 & 62.94 \\
\hspace{1em} + RAG             & 68.77\nomup & 66.27\nomup & 62.67\nomup & 62.34\nomup & 60.32\nomup & 60.64\nomdown \\
\hspace{1em} + Agent           & 61.41\nomup & 55.63\nomup & 59.74\nomdown & 61.53\nomup & 34.20\nomdown & 33.78\nomdown \\
DeepSeek-R1-14B                & 53.54 & 54.18 & 56.68 & 58.02 & 45.69 & 55.05 \\
\hspace{1em} + RAG             & 67.56\nomup & 64.21\nomup & 61.26\nomup & 61.09\nomup & 32.30\nomdown & 41.31\nomdown \\
\hspace{1em} + Agent           & 66.23\nomup & 64.07\nomup & 61.21\nomup & 61.77\nomup & 51.21\nomup & 56.78\nomup \\
\midrule

\rowcolor{black!4}
\multicolumn{7}{l}{\textbf{Summer 2025}} \\
\midrule

\multicolumn{7}{l}{\textit{Traditional Baselines}} \\
Random Forest                  & \best{65.57} & 66.48 & 55.88 & 61.63 & 46.15 & 45.65 \\
Feed-Forward NN                & \second{65.25} & \best{67.78} & 64.95 & 67.58 & 57.56 & \second{85.87} \\
L2 Logistic Regression         & 60.62 & \second{67.20} & 70.92 & 73.10 & 43.46 & 49.28 \\
KNN + Random Forest            & 62.13 & 63.29 & 57.56 & 62.88 & 46.74 & 46.74 \\
HINT                           & 64.08 & 66.86 & 56.28 & 63.89 & 46.15 & 45.65 \\
AdaBoost                       & 60.40 & 61.05 & 58.15 & 61.03 & 39.51 & 34.78 \\
XGBoost                        & 61.53 & 63.27 & 65.26 & 69.79 & 45.56 & 44.57 \\
\midrule

\multicolumn{7}{l}{\textit{Large Language Models}} \\
o3-mini                        & 59.83 & 59.46 & 72.48 & 71.93 & 54.94 & 59.78 \\
\hspace{1em} + RAG             & 59.07\nomdown & 58.98\nomdown & \second{73.15}\nomup & 72.94\nomup & 56.75\nomup & 60.87\nomup \\
\hspace{1em} + Agent           & 61.75\nomup & 62.84\nomup & \best{73.33}\nomup & \second{73.96}\nomup & \best{69.81}\sigup & \best{92.75}\sigup \\
GPT-5                          & 51.54 & 58.24 & 70.23 & 73.71 & 64.81 & 64.49 \\
\hspace{1em} + RAG             & 50.13\nomdown & 55.55\nomdown & 70.18\nomdown & 73.78\nomup & \second{68.95}\nomup & 70.05\nomup \\
Claude Opus 4.5                & 53.71 & 57.86 & 68.76 & 73.68 & 48.24 & 49.64 \\
\hspace{1em} + RAG             & 58.69\sigup & 62.72\sigup & 69.59\nomup & \best{74.10}\nomup & 48.42\nomup & 50.00\nomup \\
\midrule
\multicolumn{7}{l}{\textit{Open-Source Large Language Models}} \\
Qwen2.5-14B                    & 53.42 & 53.61 & 58.84 & 59.43 & 44.74 & 44.69 \\
\hspace{1em} + RAG             & 58.91\nomup & 58.55\nomup & 61.00\nomup & 61.02\nomup & 45.35\nomup & 44.20\nomdown \\
\hspace{1em} + Agent           & 60.60\nomup & 59.82\nomup & 55.13\nomdown & 57.85\nomdown & 32.14\nomdown & 23.68\nomdown \\
DeepSeek-R1-14B                & 52.04 & 52.25 & 57.95 & 59.65 & 41.39 & 51.21 \\
\hspace{1em} + RAG             & 58.32\nomup & 57.83\nomup & 61.48\nomup & 61.79\nomup & 32.92\nomdown & 47.71\nomdown \\
\hspace{1em} + Agent           & 57.31\nomup & 56.90\nomup & 57.59\nomdown & 58.33\nomdown & 29.63\nomdown & 21.05\nomdown \\
\bottomrule
\end{tabular}
\end{threeparttable}
\begin{tablenotes}[flushleft]
\footnotesize
\item[$\dagger$] $^{\dagger}$ Opus 4.5 knowledge cutoff date postdates Winter 2025's timestamp, possibly contaminated.
\end{tablenotes}
\caption{\small
Acc is balanced accuracy. Compared to base model, \textnomup and \textnomdown indicate nominal increases and decreases, while \textsigup denotes statistically significant improvements (95\% CI via hierarchical bootstrap sampling). Best and second-best results within each dataset and column are marked in \best{bold} and \second{underline}, respectively. All LLM results are averaged over three runs.
}
\label{tab:main_eval_results_full}
\end{table*}

\section{Traditional Baseline Implementation Details}
\label{sec:traditional_baseline_details}

\subsection{Full data feature set for tabular machine learning models}
\label{subsec:full_data_feature_set}

Table~\ref{tab:feature_schema} presents a detailed overview of the feature schema for tabular machine learning baseline models.

\begin{table}[bpt]
\centering
\footnotesize
\setlength{\tabcolsep}{6pt}
\renewcommand{\arraystretch}{1.1}
\begin{tabular}{@{}p{0.22\textwidth}@{\hskip 10pt}p{0.60\textwidth}@{\hskip 10pt}p{0.12\textwidth}@{}}
\toprule
\textbf{Feature Name} & \textbf{Description \& Examples} & \textbf{Type} \\
\midrule

\rowcolor{gray!6}
\multicolumn{3}{@{}l}{\textbf{Trial Overview}} \\
Brief title
  & Short official title of the clinical trial.
  & Textual \\
Study type
  & Trial study type. \textit{(We only examine interventional studies in this paper)}
  & Categorical \\
Trial allocation
  & How trial participants are assigned to different study arms. \textit{(We only examine randomized trials in this paper)}
  & Categorical \\
Intervention model
  & The structural design of the trial. \textit{(EXAMPLES: parallel; crossover; factorial; sequential; single group)}
  & Categorical \\
Primary purpose
  & The main intent of the trial. \textit{(EXAMPLES: treatment; prevention; device feasibility, diagnostic)}
  & Categorical \\
Trial masking
  & The blinding scheme used in the trial, indicating who does or does not know the treatment assignment. \textit{(EXAMPLES: single; double; none)}
  & Categorical \\
Enrollment count
  & The total number of participants enrolled or planned for the study.
  & Numerical (discrete) \\
Protocol arms count
  & The number of study arms defined in the protocol.
  & Numerical (discrete) \\
Protocol interventions count
  & The number of distinct interventions being studied across the trial.
  & Numerical (discrete) \\
\midrule
\rowcolor{gray!6}
\multicolumn{3}{@{}l}{\textbf{Clinical Features}} \\
Trial conditions
  & The diseases, disorders, or other medical conditions being targeted in the study.
  & Textual \\
Intervention "MeSH" terms
  & Standardized medical subject headings describing the interventions, including drug therapies, surgeries, and physical treatments.
  & Textual \\
\midrule
\rowcolor{gray!6}
\multicolumn{3}{@{}l}{\textbf{Patient Information}} \\
Eligibility criteria
  & The inclusion and exclusion criteria that determine who can or cannot participate in the study.
  & Textual \\
Sex
  & Which sex groups are eligible for enrollment. \textit{(EXAMPLES: male; female; all)}
  & Categorical \\
Minimum age
  & The youngest allowed participant age (converted to years).
  & Numerical (discrete) \\
Maximum age
  & The oldest allowed participant age (converted to years).
  & Numerical (discrete) \\
Healthy volunteers
  & Whether people without the target disease are allowed to participate.
  & Categorical (binary) \\
\midrule
\rowcolor{gray!6}
\multicolumn{3}{@{}l}{\textbf{Sponsors \& Sites}} \\
Lead sponsor name
  & The main organization responsible for leading or funding the trial. \textit{(EXAMPLES: Eli Lilly and Company; AbbVie; Merck Sharp \& Dohme LLC; Novartis Pharmaceuticals)}
  & Categorical \\
Lead sponsor class:
  & The high-level category of the lead sponsor. \textit{(EXAMPLES: industry; network; FED; NIH)}
  & Categorical \\
Collaborator classes
  & The categories of any collaborating organizations involved in the study. \textit{(EXAMPLES: FED / industry; NIH / industry)}
  & Categorical (multi) \\
Locations
  & The country or countries where the trial is being conducted. \textit{(EXAMPLES: United States; China; multiple\_countries)}
  & Categorical \\
\midrule
\rowcolor{gray!6}
\multicolumn{3}{@{}l}{\textbf{Endpoint \& Study Arms}} \\
Trial protocol arms
  & A structured description of the study arms, including which groups receive which treatments.
  & Textual \\
Endpoint details
  & A detailed description of the endpoint, explaining exactly what outcome is being evaluated.
  & Textual \\
Endpoint time frame
  & When the endpoint is assessed (converted to number of days).
  & Numerical (discrete) \\
Focal arm
  & Description of the main study arm whose outcome is being evaluated in the comparison.
  & Textual \\
Comparator arm
  & Description of the reference study arm against which the focal arm is compared.
  & Textual \\
\bottomrule
\end{tabular}
\caption{Feature schema with description, examples and feature types.}
\label{tab:feature_schema}
\end{table}

\subsection{Baseline Implementation Details and Hyperparameters}
\label{subsec:baseline_implementation_details}

All baselines share a common feature preprocessing pipeline unless noted otherwise.
\textbf{Numerical features} are used directly; missing values are imputed with the column median (KNN+RF instead uses $k$-nearest-neighbor imputation following \citet{lo2019machine}).
\textbf{Categorical features} are one-hot encoded; missing values are imputed with the most-frequent category.
\textbf{Textual features} are embedded with NV-Embed-v2~\citep{lee2024nv} and reduced to 32 dimensions via Principal Component Analysis; missing values are replaced by a zero vector.
The only exception is Logistic Regression, where it uses the full NV-Embed-v2 embeddings without PCA reduction.
For FFNN, the epoch with the best validation performance is selected for evaluation on both test benchmark datasets.

\subsection{Feature Group Importance Analysis}
\label{subsec:feature_importance}

We ran a grouped permutation-importance analysis for the Random Forest and L2 Logistic Regression baselines. For each of 15 runs per classifier (3 task types $\times$ 5 random seeds), we permuted one feature group at a time, held all remaining groups and the labels fixed, and recorded the resulting drop in macro-F1. Textual features are encoded as vectors. All embedding dimensions of a given feature were permuted jointly, so that the contribution of a feature is not scaled by its embedding dimensionality. Each reported score is the mean drop across runs, weighted by per-task sample size, aggregated over both benchmarks. A larger value corresponds to a larger drop in macro F1 under permutation of that group.

Table~\ref{tab:feature_importance} reports the union of the two models' top-10 groups, sorted by the larger of the two scores; ``---'' indicates that the group falls outside that model's top 10. For both models, the largest drops come from textual features: endpoint details, comparator arm, trial conditions, and eligibility criteria. Among non-textual features, trial locations and lead sponsor class produce the largest drops. Scores are not directly comparable across the two models, since Logistic Regression uses the full NV-Embed-v2 embeddings while Random Forest uses their 32-dimensional PCA reduction (Appendix~\ref{subsec:baseline_implementation_details} .

\begin{table}[htbp]
\centering
\small
\setlength{\tabcolsep}{16pt}
\renewcommand{\arraystretch}{1.15}
\begin{tabular}{@{}llrr@{}}
\toprule
\textbf{Feature Group} & \textbf{Type} & \textbf{Logistic Regression} & \textbf{Random Forest} \\
\midrule
Endpoint details            & Textual     & 0.0423 & 0.0247 \\
Comparator arm              & Textual     & 0.0381 & 0.0104 \\
Trial conditions            & Textual     & 0.0055 & 0.0223 \\
Locations                   & Categorical & 0.0165 & ---    \\
Eligibility criteria        & Textual     & 0.0044 & 0.0128 \\
Lead sponsor class          & Categorical & 0.0119 & 0.0025 \\
Brief title                 & Textual     & 0.0023 & 0.0115 \\
Trial protocol arms         & Textual     & ---    & 0.0086 \\
Intervention MeSH terms     & Textual     & 0.0017 & 0.0067 \\
Trial lead sponsor name           & Categorical & 0.0040 & ---    \\
Trial maximum age (years)                & Numerical   & 0.0040 & ---    \\
Focal arm                   & Textual     & ---    & 0.0034 \\
Enrollment count            & Numerical   & ---    & 0.0013 \\
\bottomrule
\end{tabular}
\caption{\small Grouped permutation importance, reported as mean drop in macro-F1
when a feature group is permuted. Rows are the union of the two models' top-10
groups, sorted by the larger of the two scores. ``---'' indicates the group falls
outside that model's top 10.}
\label{tab:feature_importance}
\end{table}

\begin{table}[ht]
\centering
\vspace{4pt}
\small
\setlength{\tabcolsep}{20pt}
\renewcommand{\arraystretch}{1.15}
\begin{tabular}{@{}llc@{}}
\toprule
\textbf{Model} & \textbf{Hyperparameter} & \textbf{End. / Sup. / Comp.} \\
\midrule
\multirow{4}{*}{\shortstack[l]{Logistic Regression\\(L2)~\citep{lo2019machine}}}
  & penalty            & \multicolumn{1}{c}{L2} \\
  & $C$ (inverse regularization strength)               & \multicolumn{1}{c}{1.0} \\
  & solver             & \multicolumn{1}{c}{lbfgs} \\
  & max\_iter          & \multicolumn{1}{c}{3000} \\
\midrule
\multirow{3}{*}{\shortstack[l]{Random Forests\\(RF)~\citep{lo2019machine}}}
  & n\_estimators       & 300 / 200 / 200 \\
  & max\_depth          & \multicolumn{1}{c}{None} \\
  & min\_samples\_leaf  & 6 / 4 / 4 \\
\midrule
\multirow{5}{*}{\shortstack[l]{KNN+RF\\~\citep{lo2019machine}}}
  & knn\_neighbors      & \multicolumn{1}{c}{5} \\
  & knn\_weights        & \multicolumn{1}{c}{uniform} \\
  & n\_estimators       & 200 / 100 / 100 \\
  & max\_depth          & \multicolumn{1}{c}{None} \\
  & min\_samples\_leaf  & \multicolumn{1}{c}{4} \\
\midrule
\multirow{5}{*}{\shortstack[l]{FFNN\\~\citep{tranchevent2019deep}}}
  & hidden\_layer\_sizes & \multicolumn{1}{c}{(512, 128)} \\
  & activation          & \multicolumn{1}{c}{ReLU} \\
  & $\alpha$            & \multicolumn{1}{c}{$1\!\times\!10^{-4}$} \\
  & learning\_rate\_init & \multicolumn{1}{c}{$1\!\times\!10^{-3}$} \\
  & max\_iter           & \multicolumn{1}{c}{200} \\
\midrule
\multirow{4}{*}{\shortstack[l]{AdaBoost\\~\citep{fan2020application}}}
  & n\_estimators        & 300 / 300 / 200 \\
  & learning\_rate       & \multicolumn{1}{c}{1.0} \\
  & base\_max\_depth     & \multicolumn{1}{c}{2} \\
  & base\_min\_samples\_leaf & \multicolumn{1}{c}{2} \\
\midrule
\multirow{3}{*}{\shortstack[l]{XGBoost\\~\citep{siah2021predicting}}}
  & n\_estimators       & \multicolumn{1}{c}{300} \\
  & learning\_rate      & 0.1 / 1.0 / 1.0 \\
  & max\_depth          & \multicolumn{1}{c}{12} \\
\bottomrule
\end{tabular}
\caption{\small Hyperparameters for each baseline. Where a single value is listed, the same setting is used across all three task types (Endpoint / Superiority / Comparative Effect). Otherwise, task-specific values are shown separated by ``/''.}
\label{tab:hyperparams}
\end{table}

\section{Pipelines}
\label{subsec:pipelines}

\subsection{Decontamination Pipeline Implementation Detail}
\label{subsec:decontamination_detail}

We provide implementation details for the decontamination pipeline summarized in Algorithm~\ref{alg:decontamination}. In Algorithm~\ref{alg:decontamination} we use a simplified notation $\textsc{Search}(query, [start, end], mode)$. In fact, we used this notation $\textsc{Search}(query, [start, end], mode, k)$, which retrieves up to $k$ documents about a clinical trial that report results, subject to the publication date constraint $[start, end]$.

\begin{figure}[htbp]
    \centering
    \includegraphics[width=\linewidth]{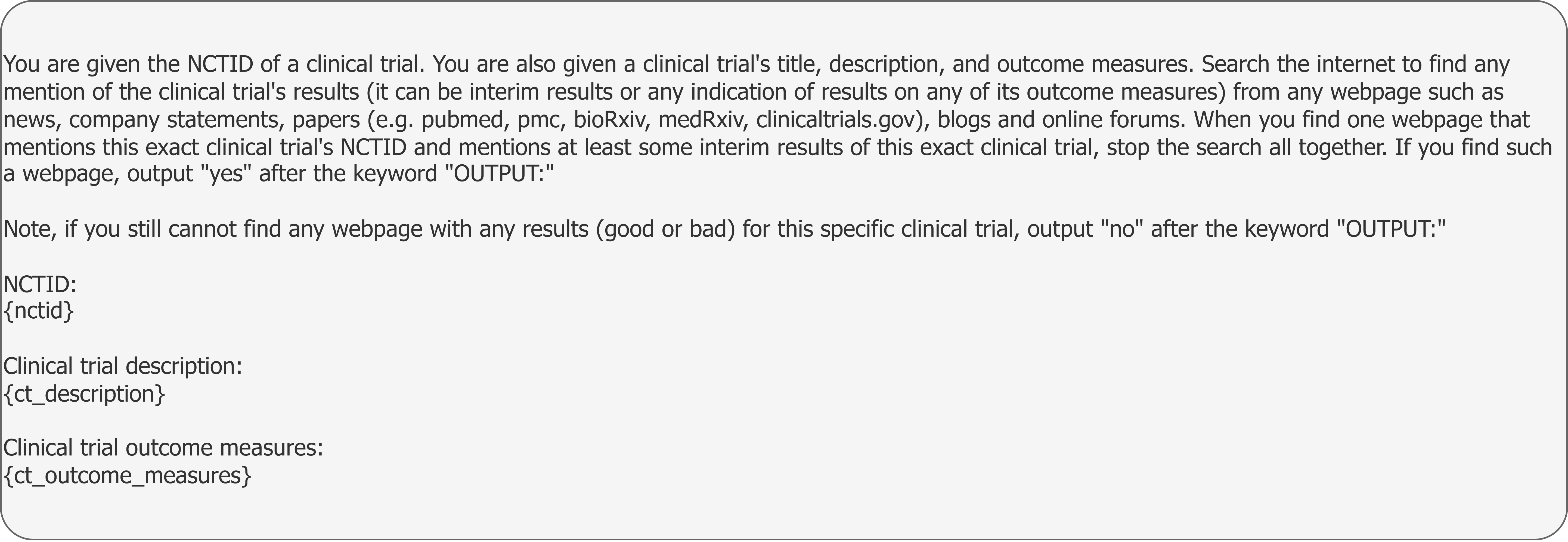}
    \caption{Prompt for the initial screening pass, used to determine whether any results exist on the web for a given trial.}
    \label{fig:initial_check_result_prompt}
\end{figure}

\paragraph{Initial Screening Pass.}
As shown in Algorithm~\ref{alg:decontamination}, the first step of the pipeline is a quick screening pass over the full initial trial pool $\mathcal{P}$, which typically contains over 10,000 trials. For each trial $t$, we run $\textsc{Search}(t, [-\infty, +\infty], \text{GPT-5.2}, 1)$, which terminates as soon as a single document reporting results is found. Trials for which no result is found anywhere on the web are immediately discarded, since they cannot serve as valid evaluation targets. This step reduces the candidate set from approximately 10,000 trials to approximately 2,000, significantly lowering the cost of the more intensive searches that follow. The prompt used for this initial screening is shown in Figure~\ref{fig:initial_check_result_prompt}.

\begin{figure}[htbp]
    \centering
    \includegraphics[width=\linewidth]{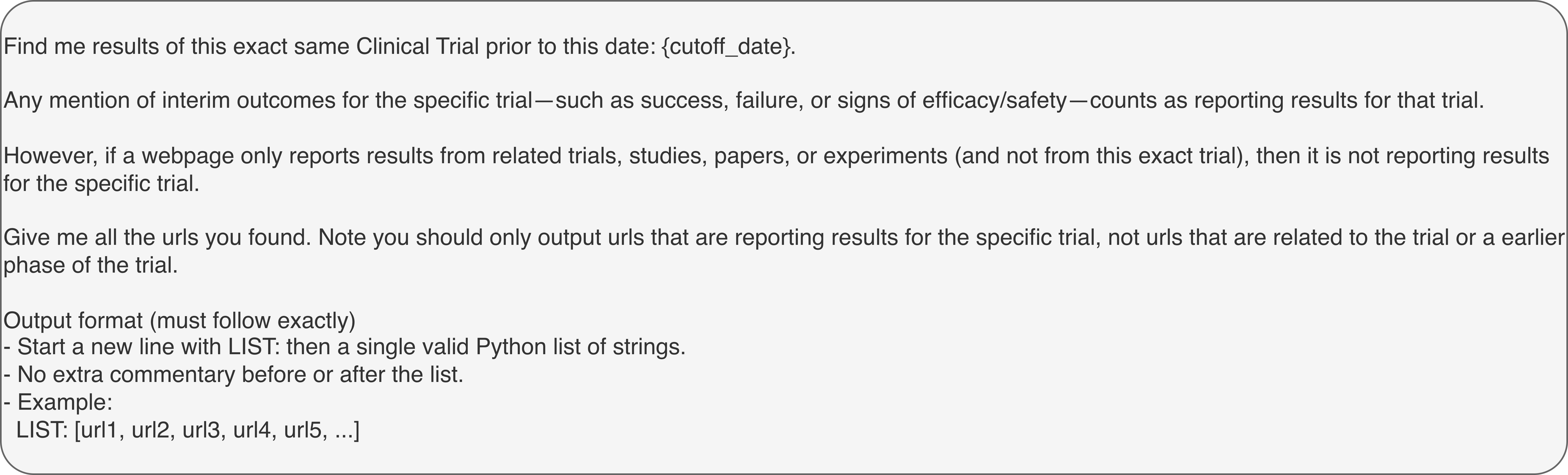}
    \caption{\small Prompt for LLM web search before the cutoff date.}
    \label{fig:LLM_websearch_before_cutoff_prompt}
\end{figure}

\begin{figure}[htbp]
    \centering
    \includegraphics[width=\linewidth]{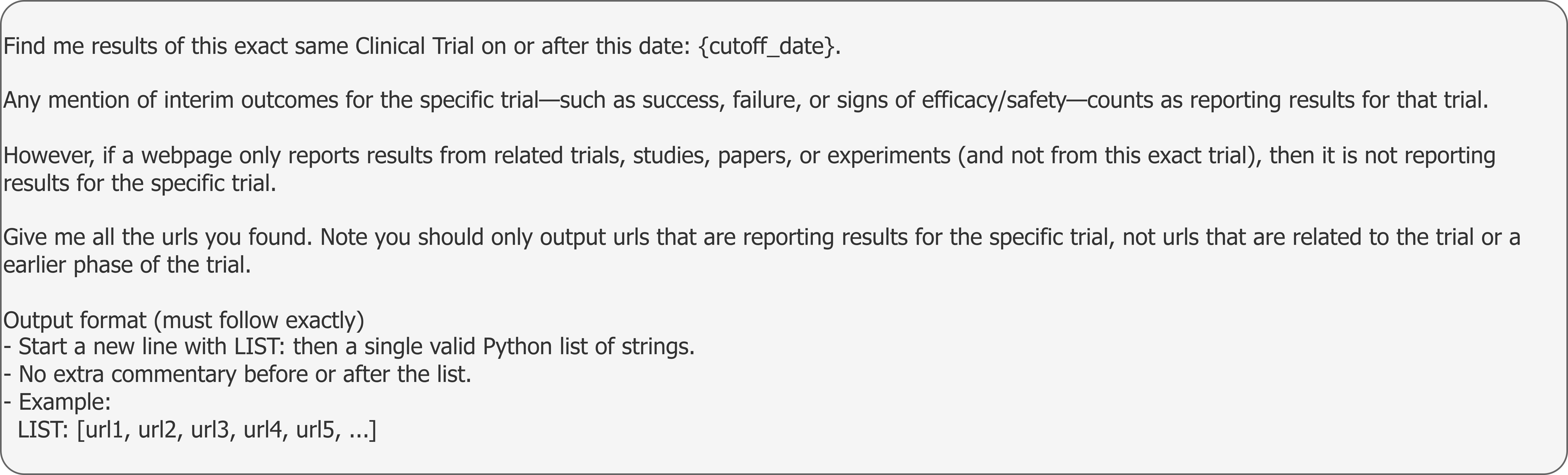}
    \caption{\small Prompt for LLM web search after the cutoff date.}
    \label{fig:LLM_websearch_after_cutoff_prompt}
\end{figure}

\begin{figure}[htbp]
    \centering
    \includegraphics[width=\linewidth]{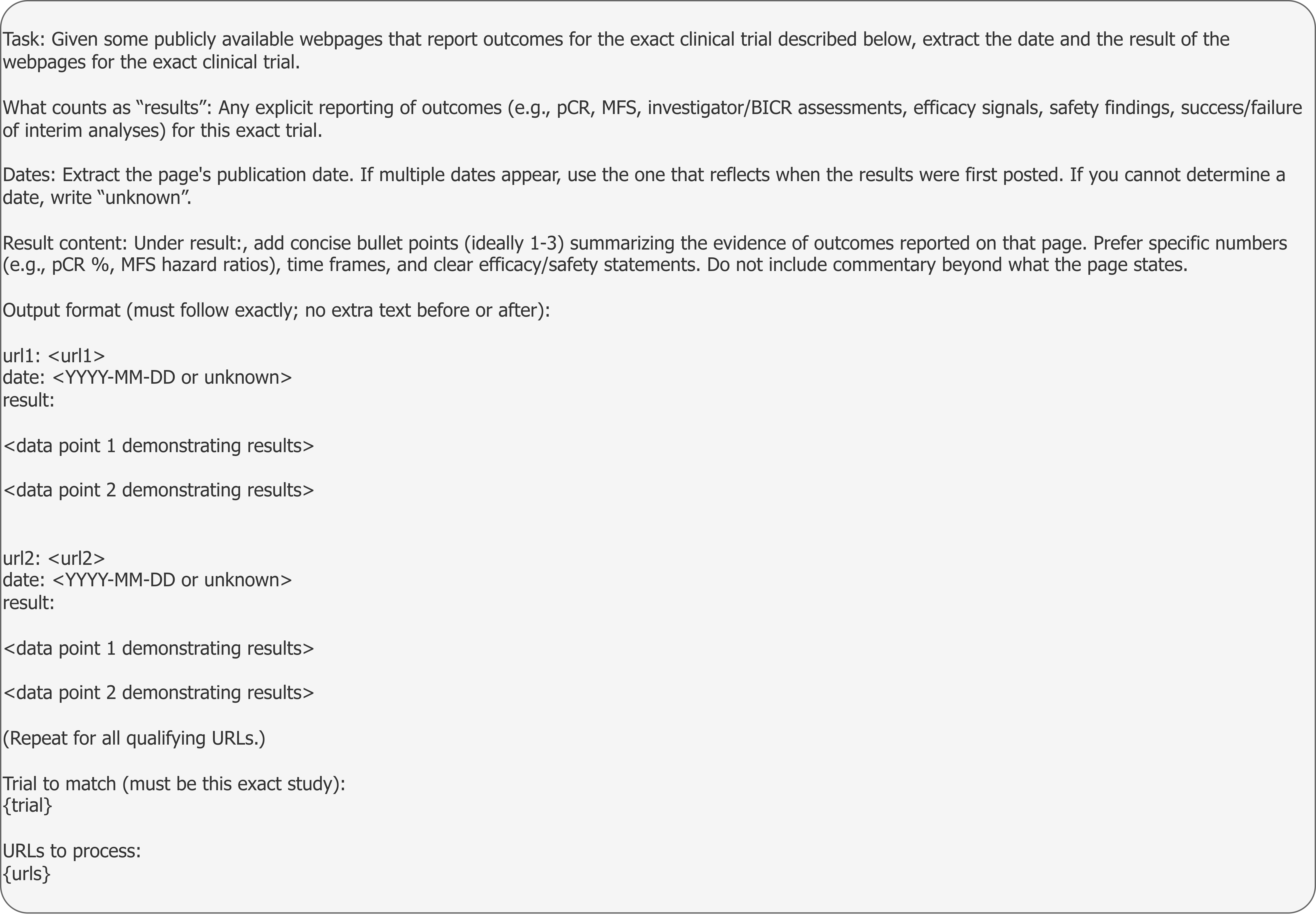}
    \caption{\small Prompt for extracting publication dates and result summaries from URLs discovered by LLM web search.}
    \label{fig:LLM_websearch_result_extraction_prompt}
\end{figure}

\paragraph{LLM Mode Search.}
When $mode$ is an LLM (GPT-5.2 or Gemini~3), search is performed through the corresponding model's web search capability, as described in the main text. We implement this in two rounds. For each trial, we flatten the protocol section (excluding eligibility criteria and any date containing fields) into a text representation. In Round~1, we ask the model to perform a grounded web search and return a list of URLs that report results for that exact trial (not related trials or earlier phases). The prompts used for the before-cutoff and after-cutoff searches are shown in Figure~\ref{fig:LLM_websearch_before_cutoff_prompt} and Figure~\ref{fig:LLM_websearch_after_cutoff_prompt}, respectively. We normalize the raw LLM output into clean URL lists, handling varied formats such as Python literals, JSON, and raw text with URL extraction. In Round~2, we send the discovered URLs back to the model with a second prompt asking it to visit each URL and extract the publication date and concise result summaries (e.g., pCR rates, hazard ratios, efficacy/safety findings). The prompt used for this extraction step is shown in Figure~\ref{fig:LLM_websearch_result_extraction_prompt}. We parse this second round output using regex to segment multi-URL responses into structured dictionaries mapping each URL to its date and extracted results.

As specified in Algorithm~\ref{alg:decontamination}, we run LLM mode searches in four configurations: $\textsc{Search}(\mathrm{info}, [-\infty, c), \text{GPT-5.2}, \infty)$, $\textsc{Search}(\mathrm{info}, [-\infty, c), \text{Gemini 3}, \infty)$, $\textsc{Search}(\mathrm{info}, [c, +\infty), \text{GPT-5.2}, \infty)$, and $\textsc{Search}(\mathrm{info}, [c, +\infty), \text{Gemini 3}, \infty)$. If either model finds results before the cutoff date $c$, the trial is removed. If neither model finds any results at all, the trial is also removed, since the initial screening pass may have been a false positive.

\begin{figure}[htbp]
    \centering
    \includegraphics[width=\linewidth]{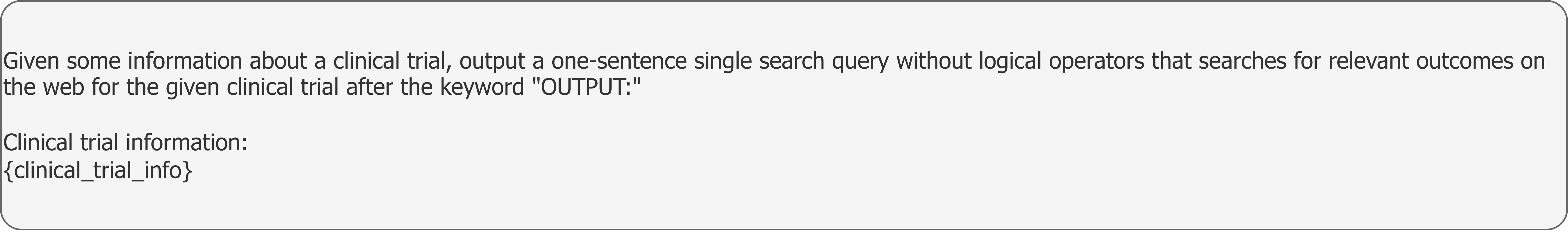}
    \caption{\small Prompt for rewriting trial information into search queries for the Brave Search API.}
    \label{fig:llm_query_rewrite_prompt}
\end{figure}

\paragraph{Brave Mode Search.}
When $mode$ is Brave, as described in the main text, the Brave Search API returns candidate URLs, webpages are fetched with Python Requests or ZenRows, and downloadable files are converted to text with DeepSeek OCR. We use three query strategies derived from the trial's information $\mathrm{info} = \{\text{NCT ID, aliases, title, description}\}$: (1) the trial's brief title, (2) its NCT ID, and (3) an LLM-rewritten query generated by GPT-5 that incorporates identifiers, tags, eligibility criteria, and other metadata from the trial protocol. The prompt used for query rewriting is shown in Figure~\ref{fig:llm_query_rewrite_prompt}. Each query type is run both with and without a freshness date filter, where the freshness parameter restricts results based on the trial's study start date and the earliest available completion or results submission date. For each query configuration, we fetch up to 100 results, and all searches are executed in parallel using multithreading. The raw search results from all query and freshness combinations are then collected. For each returned URL, we determine its type to select the appropriate content retrieval method. If the URL points to a known paper repository (e.g., PubMed, PMC, or a preprint server), we use the corresponding database API to retrieve the paper's full text and publication date directly. If the URL points to a downloadable PDF, we download the file and convert it to text using DeepSeek-OCR2, deployed on our local H100 server. For all other URLs, we use the Python Requests library along with the ZenRows API to scrape the full webpage content. The scraped and converted texts are then stored in a database for downstream processing and verification.

\paragraph{Database Mode Search.}
When $mode$ is Database, we use the trial's NCT ID to search PubMed, PMC, BioRxiv, and MedRxiv for matching articles, as described in the main text.

\paragraph{Text Extraction from Scraped Webpages.}
For Brave mode and Database mode searches, all retrieved texts, including webpages, downloadable files, and database articles, undergo our customized processing pipeline. As noted in the main text, scraped webpages are processed with a customized text extraction system derived from trafilatura, designed for long pages and for isolating the main article content while removing HTML, sidebars, and other articles' text. We provide further details on this system below.

We parse each page's raw HTML in parallel using multiprocessing. We first remove all scripting, styling, and non-rendered elements, then perform a depth-first traversal of the DOM tree, converting the HTML into a flat list of text and path tuples while preserving structural semantics such as headings, paragraphs, and lists in Markdown format. We then apply a series of domain-specific and general-purpose filters to remove irrelevant content. For ResearchGate pages, we truncate everything after the Citations or References section to remove recommended papers. For ichgcp.net, we remove ``similar clinical trials'' sections, and for other domains we detect and strip ``similar trials'' or ``similar articles'' blocks. We also remove PubMed tail sections such as conflict of interest statements and comment blocks, site-specific footers (e.g., Mayo Clinic), and newsletter signup sections that appear after references. A broad marketing slogan filter detects over 50 common promotional patterns such as ``You Might Also Like,'' ``Related Posts,'' ``Recommended Reading,'' ``Subscribe,'' and similar calls to action, removing them only when they appear in the latter half of the page to avoid false positives on actual content. For WithPower pages, we strip ``Other People Viewed'' sections, and for Academia pages, we remove ``Related papers'' blocks. Finally, we detect reference, bibliography, and conflict of interest sections using extensive pattern matching that handles numerous edge cases where section headers may be split across multiple HTML elements (e.g., ``R'' + ``eferences'', ``Re'' + ``ferences'', or even character by character), as well as non-English variants. When exactly one such section boundary is found, we truncate the content there; when multiple candidates exist, we flag them for further disambiguation.

\paragraph{Publication Date Extraction.}
As described in the main text, for Brave mode and Database mode searches, all retrieved texts undergo our customized pipeline for publication date extraction to verify that each document falls within the required date range $[start, end]$. We run a multi-strategy prioritized pipeline in parallel across all scraped pages to determine each webpage's publication date as accurately as possible. We first skip known problematic domains and pages that failed scraping. For preprint servers such as bioRxiv and medRxiv, we use a specialized parser that extracts the posting date from their unique HTML layout. For YouTube pages, we parse the embedded JavaScript data object to retrieve the video's publication date. For all other pages, we apply a cascading series of extraction strategies in priority order: (1)~structured data embedded in the page's JSON-LD blocks, (2)~high-priority Open Graph and article metadata tags, (3)~semantic publish date attributes on \texttt{time} and other elements, (4)~domain-specific handlers such as definition list pairs for ScienceDaily and dedicated publish date containers, (5)~citation metadata commonly used by academic publishers, (6)~dates embedded within inline JavaScript variables, (7)~semantic date spans with consistency checks for uniqueness, (8)~the earliest date found across all time-related elements on the page, and (9)~as a last resort, visible text scanning within the main content area for date patterns near keywords like ``Published,'' ``Posted,'' or ``Updated.'' When multiple strategies yield dates, we take the earliest one to be conservative. All raw date strings are normalized through a robust converter that handles over a dozen formats including ISO~8601 datetimes, year-first and day-first numeric dates, two-digit year pivoting, compact date strings, month name variants in multiple orderings, year-plus-month only, and year only, with validation for month and day ranges including leap years. Dates with years before 1990 or after the current year are rejected as invalid.

\paragraph{Two-Round GPT-5 Verification.}
As described in the main text, for Brave mode and Database mode searches, after publication date filtering, all documents undergo two rounds of GPT-5 verification: first, to confirm that the document refers to the same clinical trial; and second, to confirm that it reports some form of results for that trial. We provide further details on each round below.

\begin{figure}[ht]
    \centering
    \includegraphics[width=\linewidth]{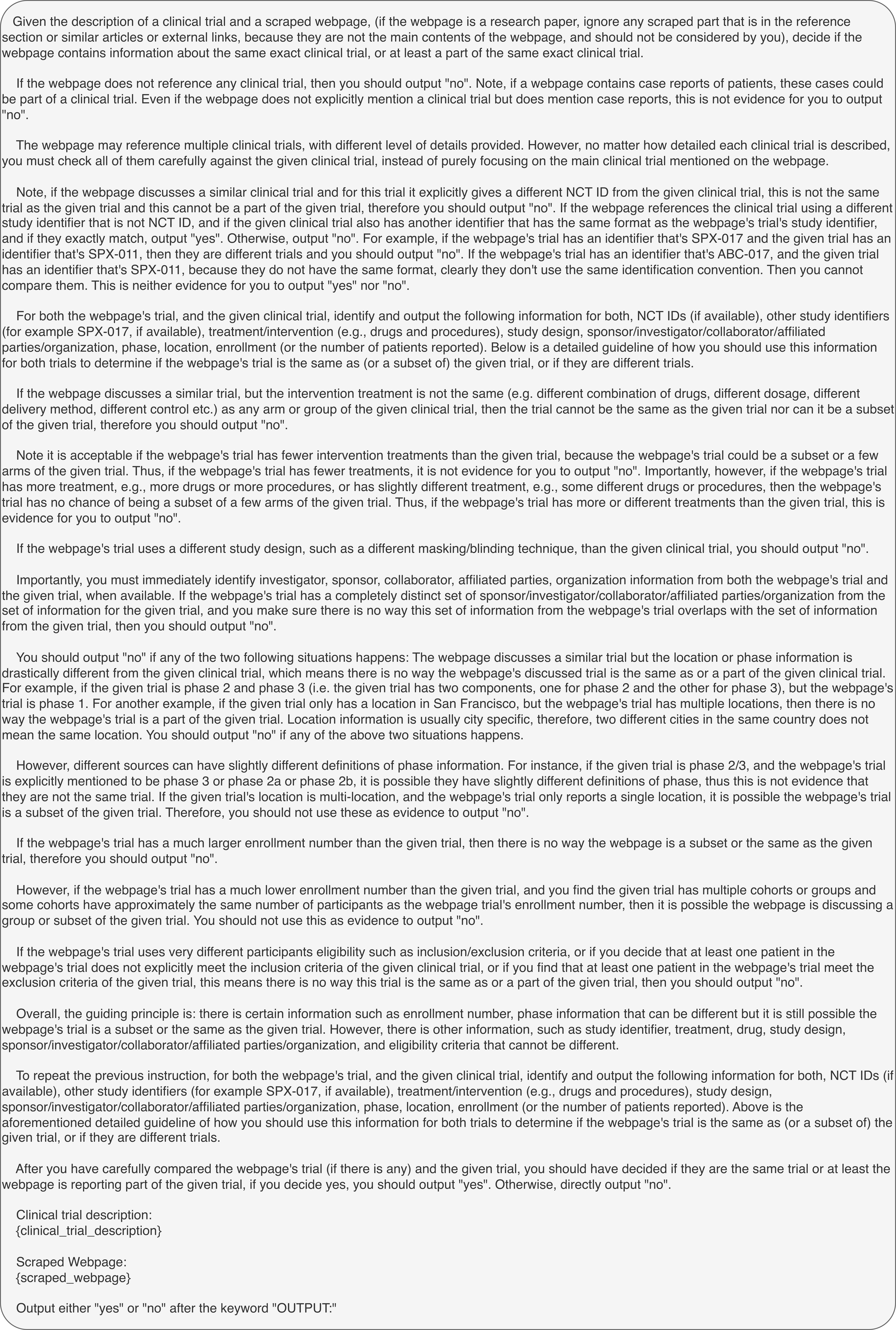}
    \caption{\small Prompt for Round~1 of GPT-5 verification, used to determine whether a document discusses the same clinical trial.}
    \label{fig:clinical_trial_identification_prompt}
\end{figure}

In Round~1 (trial matching), for each trial and document pair, we strip all date-related fields from the trial's protocol metadata to prevent the model from using temporal cues when making its judgment. We then ask GPT-5 to determine whether the document discusses the same exact clinical trial, or at least a subset of it. The prompt used for this step is shown in Figure~\ref{fig:clinical_trial_identification_prompt}. A key design choice is that we do not rely solely on NCT ID matching, since many documents, particularly news articles, press releases, and research papers, discuss clinical trials without ever mentioning a registry identifier. Instead, we instruct the model to systematically compare both sides across multiple dimensions: registry IDs and other study identifiers, intervention treatments including specific drugs and procedures, study design and masking techniques, sponsor and investigator organizations, trial phase, geographic locations, enrollment numbers, and participant eligibility criteria. By triangulating across all of these fields, the model can confidently identify a trial even when no NCT ID is present in the document, using the combination of matching drug names, sponsor organizations, study design, and enrollment characteristics as converging evidence. We enforce asymmetric matching rules: certain fields such as identifiers, treatments, study design, and sponsors must match exactly or be a strict subset, while other fields such as enrollment and phase are allowed to differ if the document could plausibly be reporting on a subset or arm of the given trial. For example, a document reporting fewer treatments or a smaller enrollment is acceptable as a potential subset, but a document with additional or different drugs, a conflicting NCT ID, or a completely distinct set of sponsors is grounds for rejection.

\begin{figure}[ht]
    \centering
    \includegraphics[width=\linewidth]{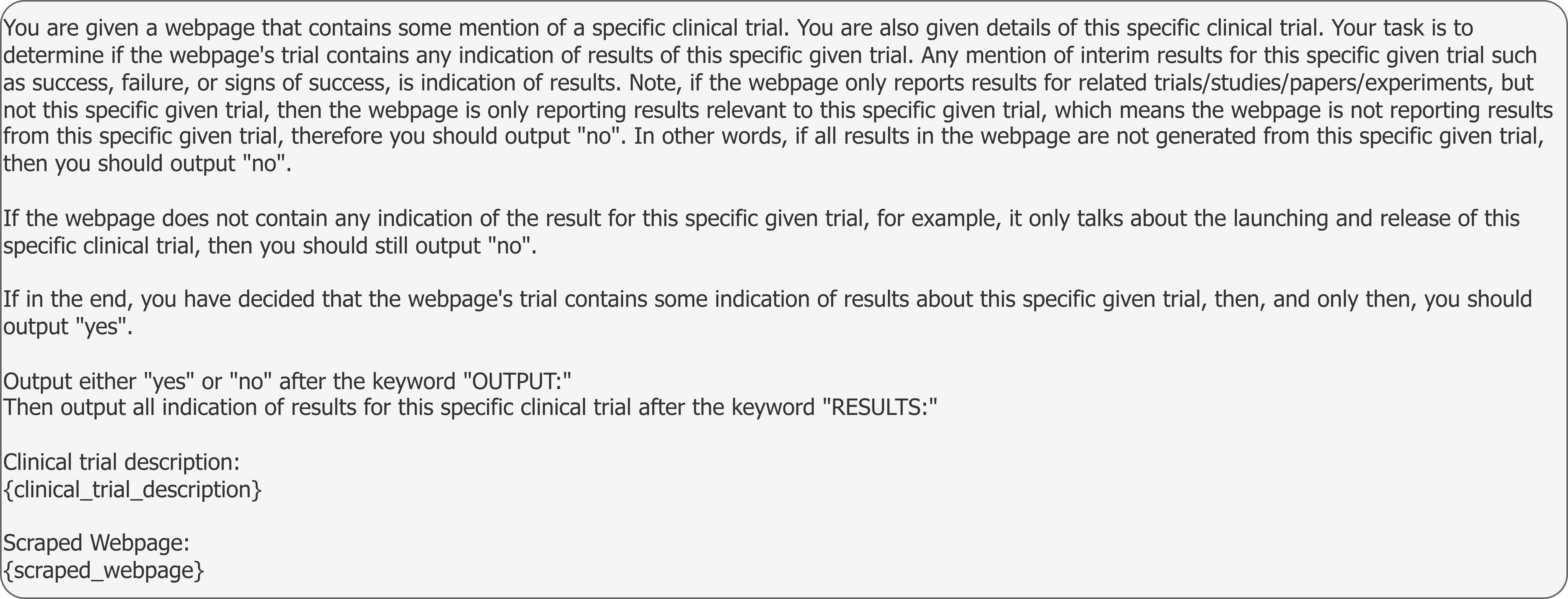}
    \caption{\small Prompt for Round~2 of GPT-5 verification, used to determine whether a document reports results for the matched trial.}
    \label{fig:result_verification_prompt}
\end{figure}

In Round~2 (results verification), only pairs that the model identifies as matching in Round~1 proceed. We ask GPT-5 again with the same trial and document pair, this time to determine whether the document reports some form of results for that trial. The prompt used for this step is shown in Figure~\ref{fig:result_verification_prompt}. We explicitly define ``results'' as any mention of outcomes such as success, failure, or signs of efficacy or safety, and instruct the model to reject cases where the document only reports results from related trials, studies, or experiments rather than from this exact trial, or where it merely discusses the launch or initiation of the trial without reporting outcomes. This two-round approach ensures that we first confirm identity between the document and the trial before checking for result contamination, reducing false positives from documents that discuss similar but distinct trials.

\subsection{Answer Generation Pipeline Implementation Detail}
\label{subsec:answer_generation_detail}

\begin{figure}[ht]
    \centering
    \includegraphics[width=\linewidth]{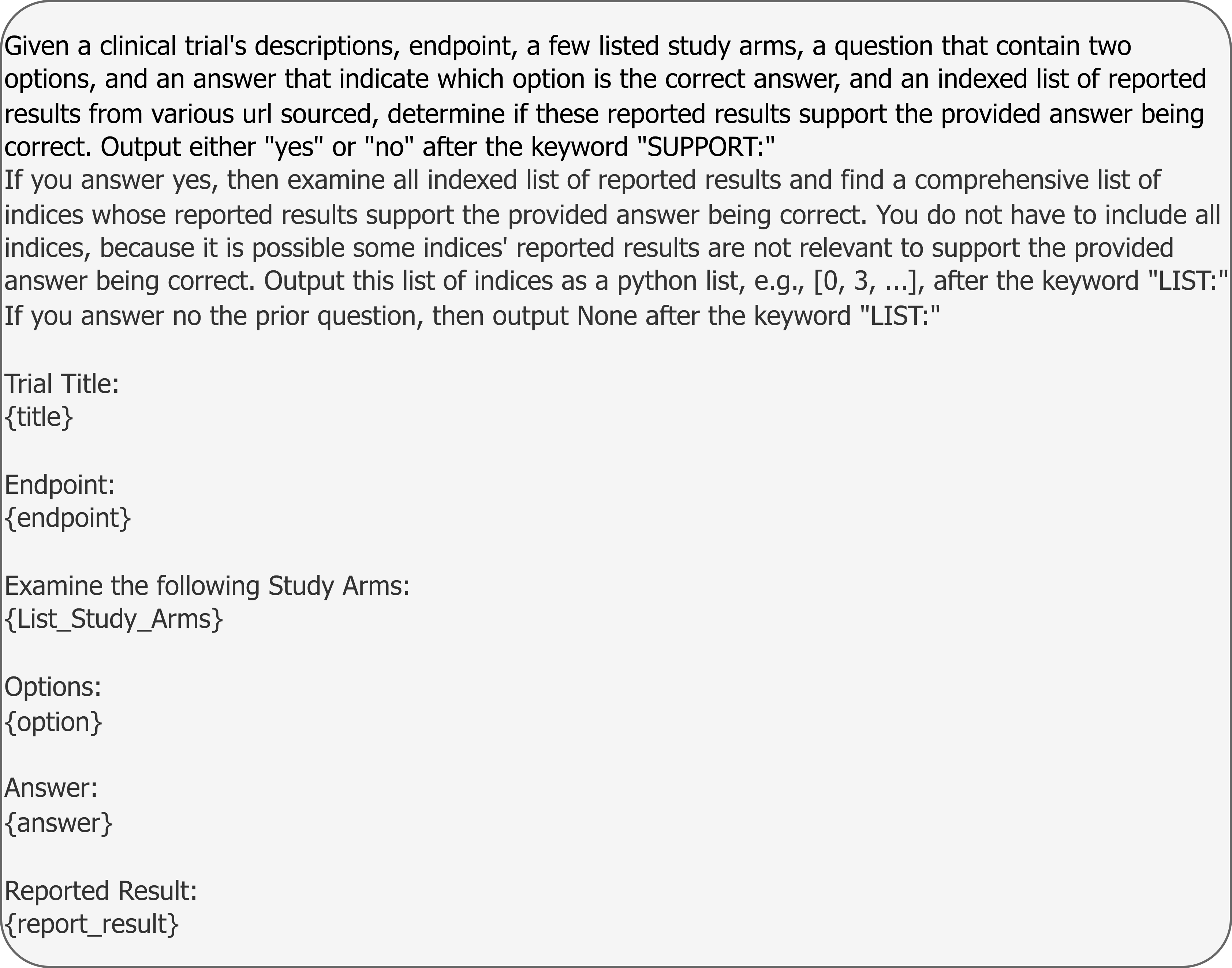}
    \caption{\small Prompt for answer verification for Endpoint/Superiority Questions.}
    \label{fig:answer_verification_endpoint_superiority}
\end{figure}

\paragraph{Answer Verification for Endpoint/Superiority Questions.}
After automatic question generation, we perform an additional LLM-based verification step to ensure the correctness of each answer label against the reported results identified earlier. Specifically, we prompt \texttt{GPT-5.4-2026-03-05} with the trial title, the endpoint under evaluation, the relevant study arms, the two answer options, the selected answer, and an indexed list of reported results collected from online sources. The model is asked to judge whether the reported results support the provided answer, returning a binary \texttt{yes} or \texttt{no} decision. Cases flagged as unsupported are removed from the benchmark. The full answer verification prompt is provided in Figure \ref{fig:answer_verification_endpoint_superiority}.

\paragraph{Answer Verification for Comparative Effect Questions.}

\begin{figure}[ht]
    \centering
    \includegraphics[width=\linewidth]{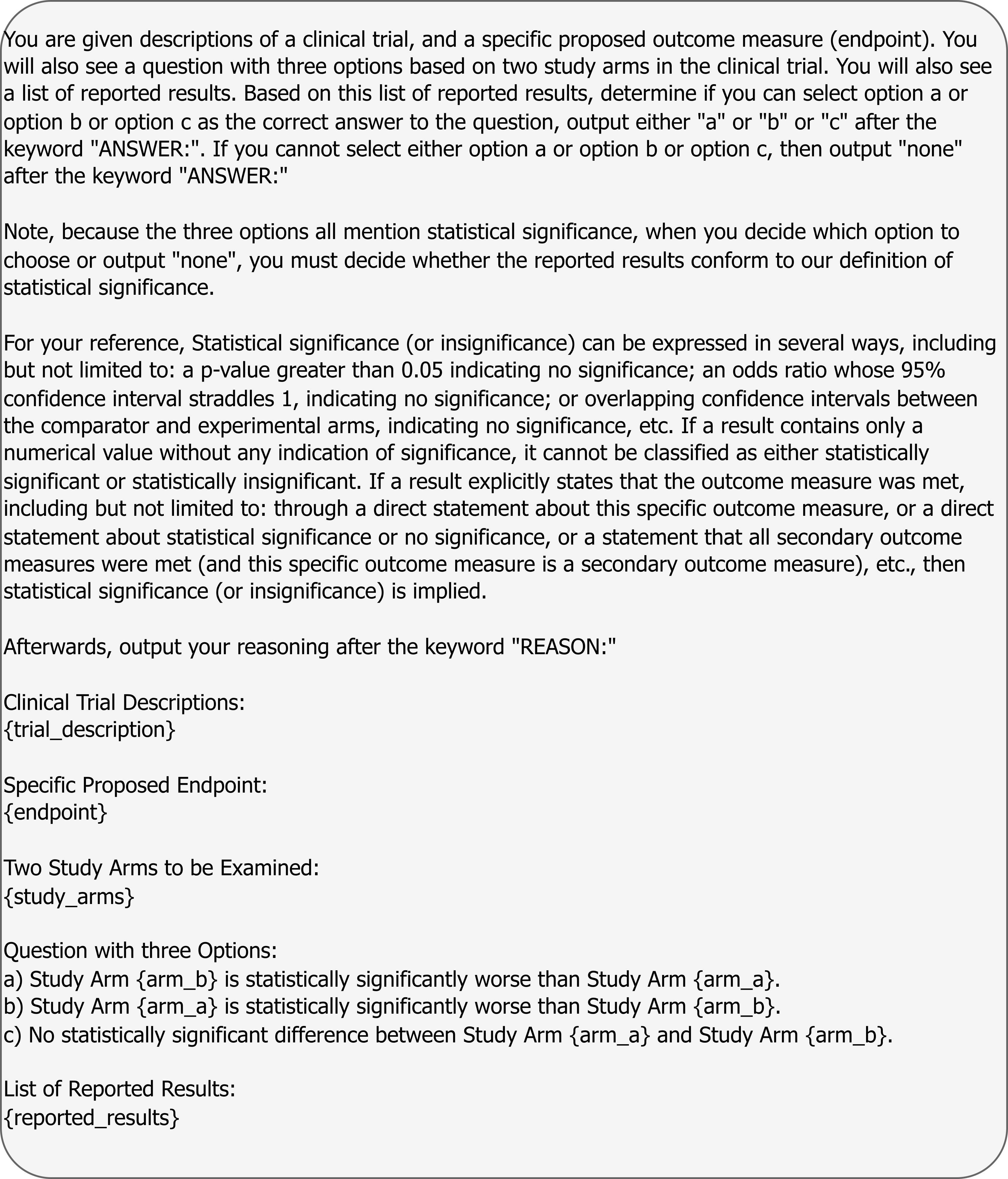}
    \caption{\small Prompt for answer verification for Comparative Effect Questions.}
    \label{fig:comparative_effect_answer_verification}
\end{figure}

For Comparative Effect questions, which involve a three-way classification between two study arms, we used a separate verification prompt. We provide \texttt{GPT-5.4-2026-03-05} with the clinical trial description, the specific endpoint under evaluation, the two study arms being compared, and the list of reported results. The model is asked to determine which of three options is correct: (a) the first arm is statistically significantly worse, (b) the second arm is statistically significantly worse, or (c) no statistical significant difference exists between the two arms. Because all three options hinge on statistical significance, the prompt includes explicit guidance on how to interpret significance from reported results. For example, a p-value exceeding 0.05, an odds ratio whose 95\% confidence interval crosses 1, or overlapping confidence intervals between arms all indicate non-significance. Conversely, if a result explicitly states that the outcome measure was met or provides a direct statement about significance, this is taken as sufficient evidence. Results that contains only numerical values without any indication of significance are treated as inconclusive. If the model cannot select an correct option and outputs \texttt{None}, the corresponding question is discarded. The full verification prompt is provided in Figure \ref{fig:comparative_effect_answer_verification}.
\clearpage

\section{Dataset Generation Prompts}
\label{subsec:dataset_generation_prompts}
This section collects all prompt templates used throughout the CT Open pipeline, including those for decontamination search, trial matching, results verification, and answer generation.



\clearpage
\section{Evaluation Prompts}
\label{subsec:evaluation_prompts}
\begin{figure}[ht]
    \centering
    \includegraphics[width=\linewidth]{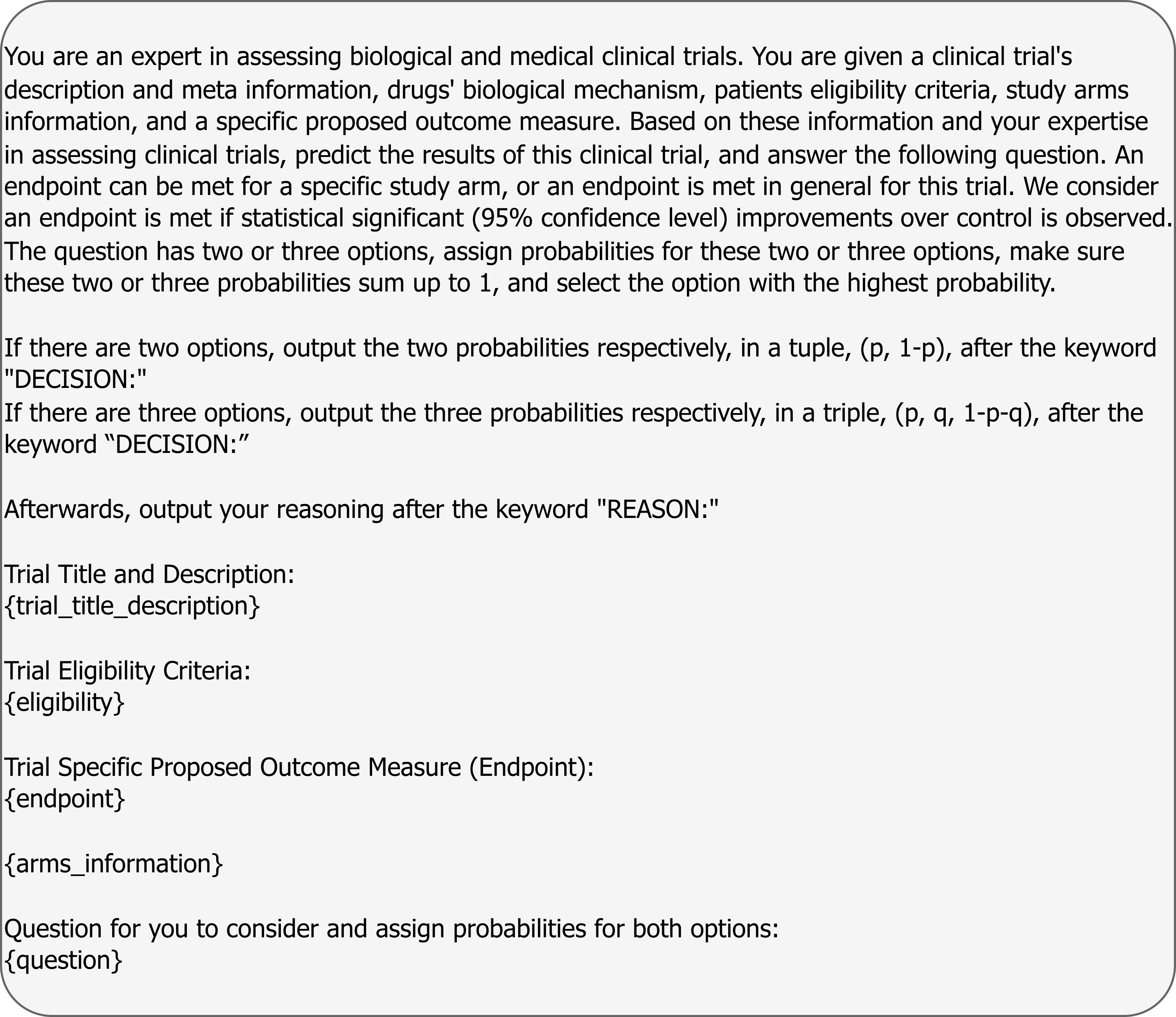}
    \caption{\small Prompt for Prompt-only LLM Evaluation.}
    \label{fig:baseline_evaluation_prompt}
\end{figure}

\begin{figure}[ht]
    \centering
    \includegraphics[width=\linewidth]{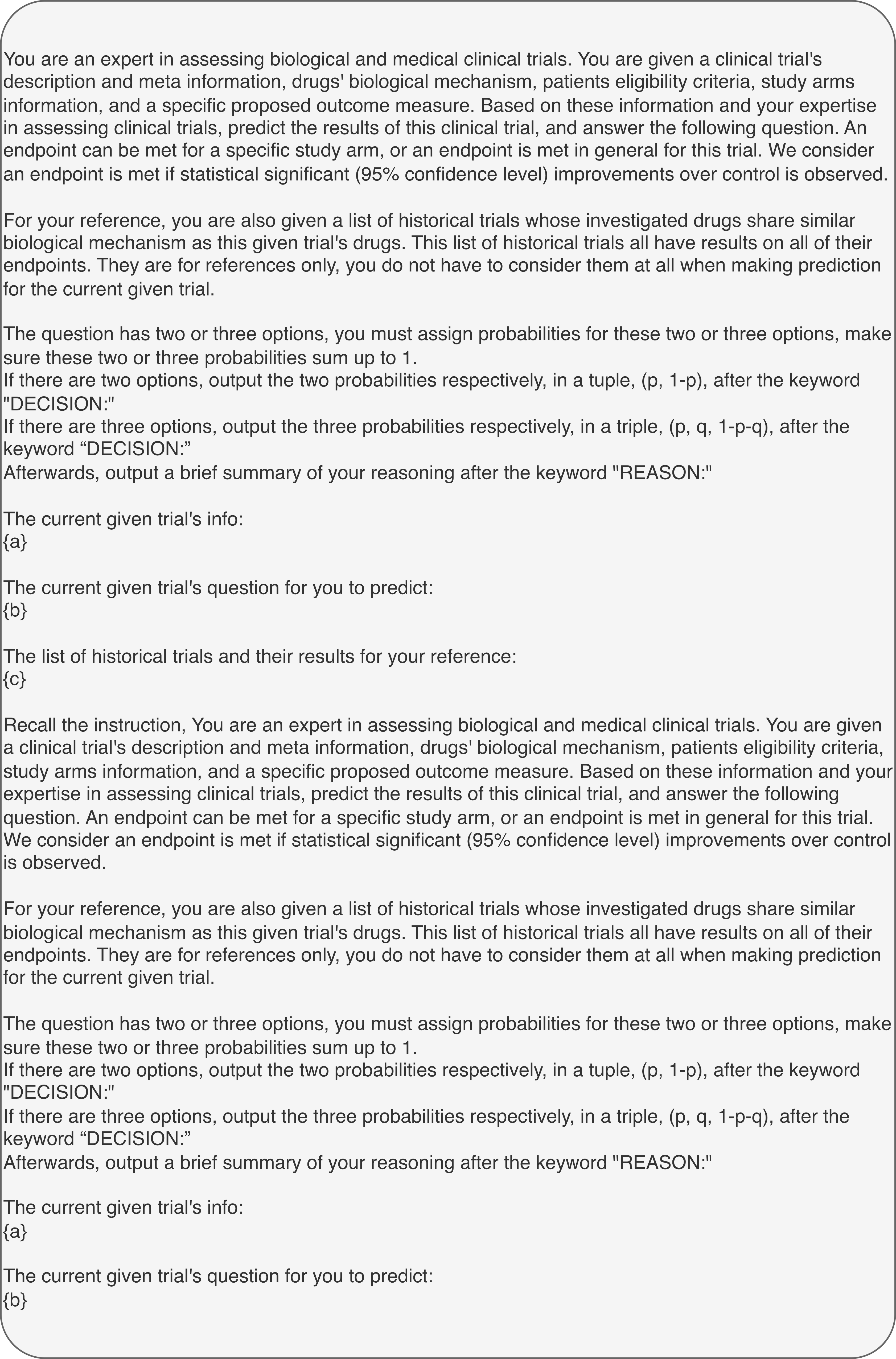}
    \caption{\small Prompt for RAG Evaluation for datapoints with Matched Historical Similar Trials.}
    \label{fig:RAG_similar_evaluation_prompt}
\end{figure}

\begin{figure}[ht]
    \centering
    \includegraphics[width=0.95\linewidth]{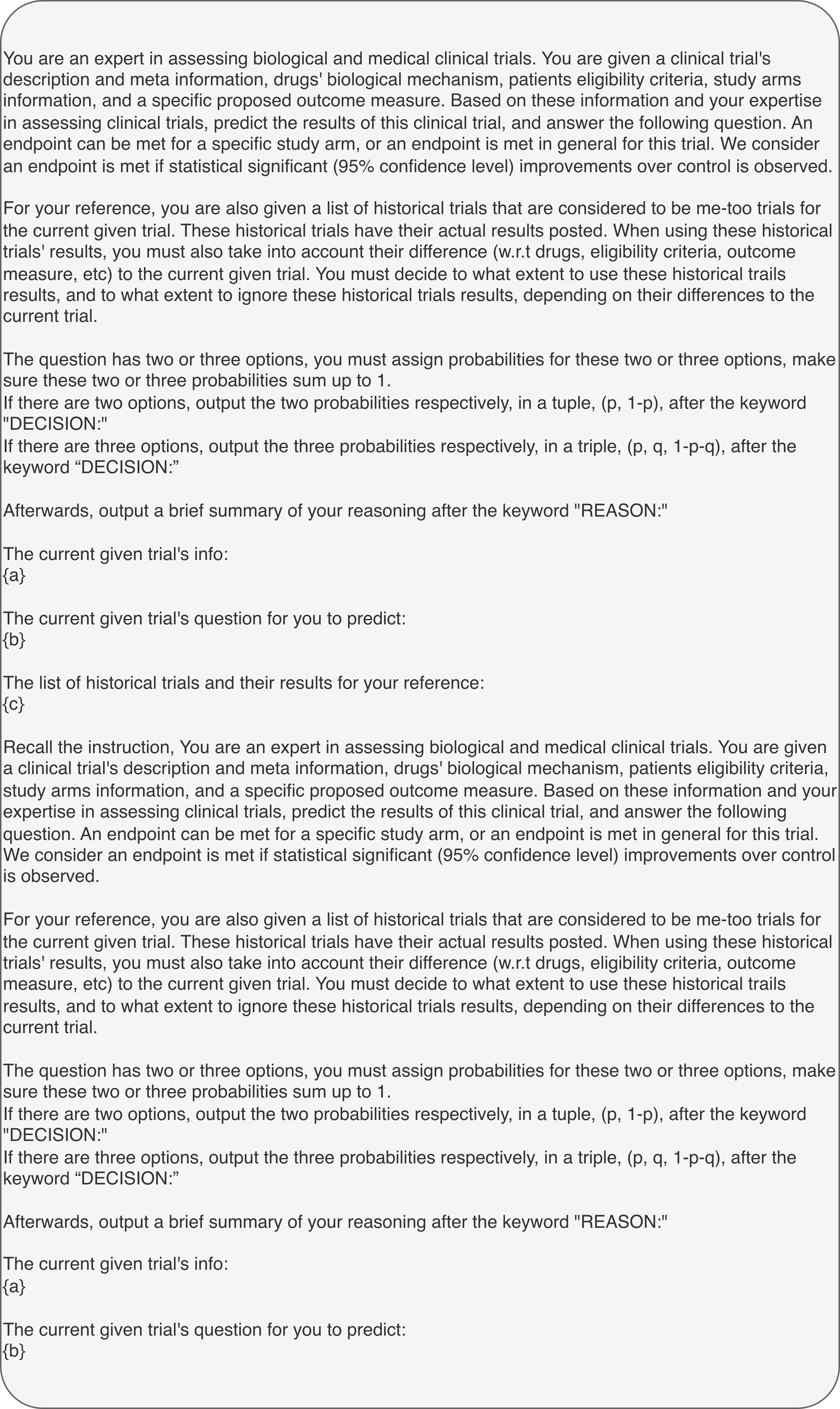}
    \caption{\small Prompt for RAG Evaluation for datapoints with Matched Historical Me-Too Trials.}
    \label{fig:RAG_metoo_evaluation_prompt}
\end{figure}

\begin{figure}[ht]
    \centering
    \includegraphics[width=\linewidth]{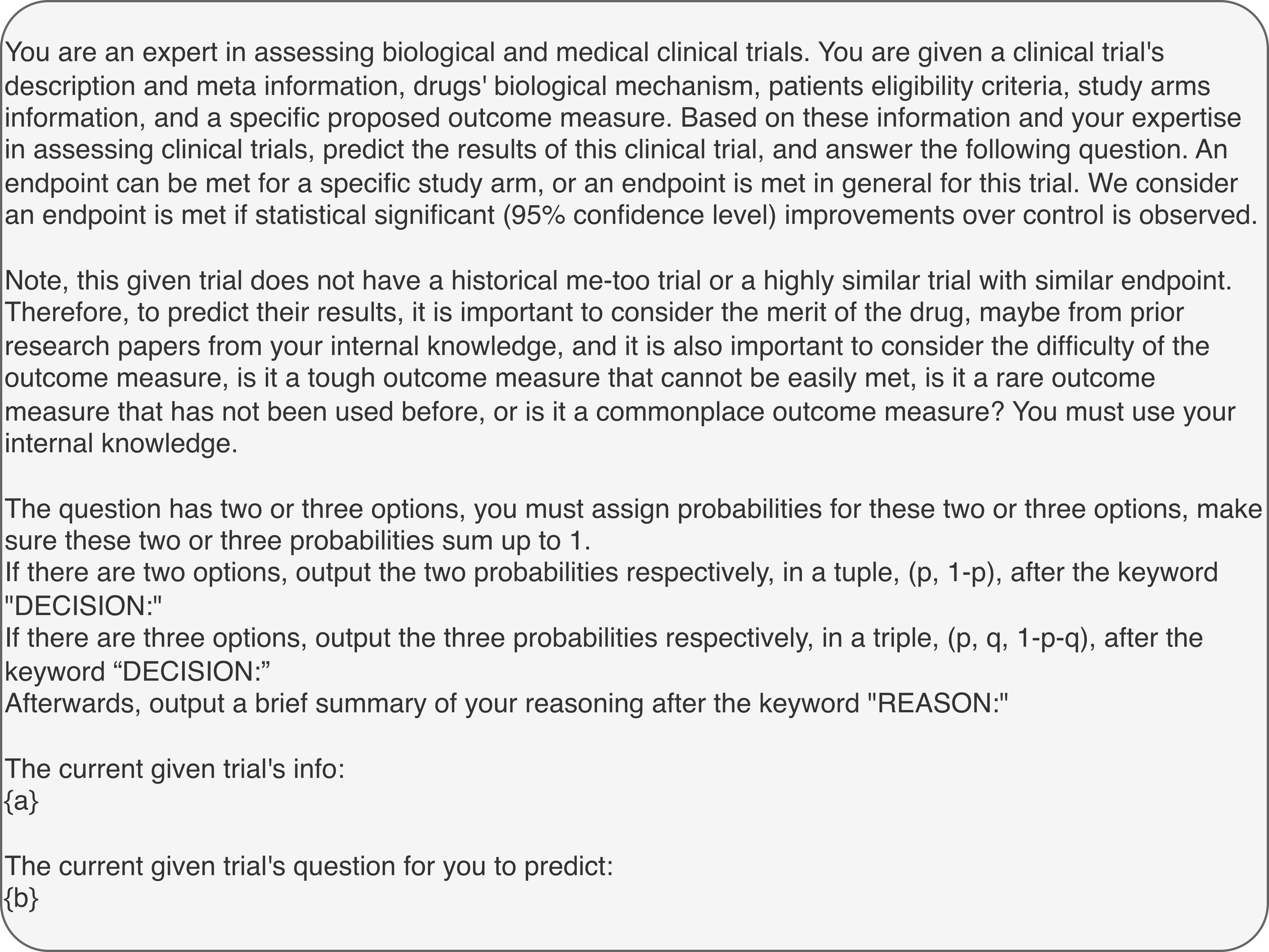}
    \caption{\small Prompt for RAG Evaluation for datapoints without Matched Historical Trials.}
    \label{fig:RAG_no_match_evaluation_prompt}
\end{figure}

\begin{figure}[ht]
    \centering
    \includegraphics[width=0.93\linewidth]{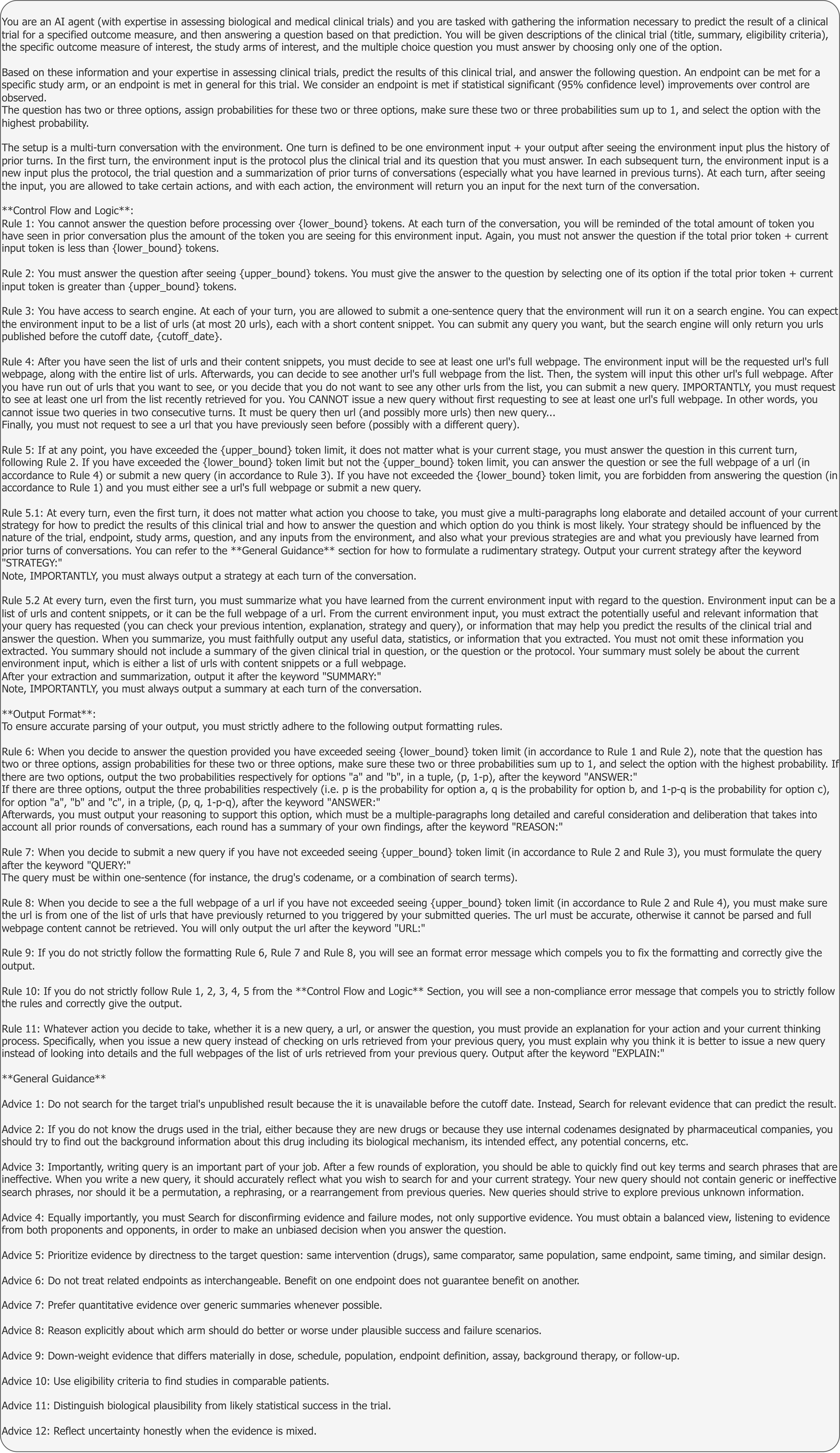}
    \caption{\small Protocol used during the agent evaluation pipeline.}
    \label{fig:agent_evaluation_protocol}
\end{figure}

This section describes the prompts used to evaluate different models. Baseline evaluation prompt for prompt-only LLMs is provided in Figure \ref{fig:baseline_evaluation_prompt}. RAG evaluation prompts for datapoints with historical similar trials, historical me-too trials, and without historical similar trials are provided in Figures \ref{fig:RAG_similar_evaluation_prompt}, \ref{fig:RAG_metoo_evaluation_prompt} and \ref{fig:RAG_no_match_evaluation_prompt}. Protocol used in agent pipeline are in Figure \ref{fig:agent_evaluation_protocol}.

\newpage

\section{Question Classes, Label Distribution, and Example Questions}
\label{sec:example_questions}

\paragraph{Answer Options.}
The meaning of each answer option depends on the question class. Throughout, ``statistically significant'' refers to the 95\% confidence level.

For the two binary classes, Endpoint and Superiority, Option~A is the positive outcome and Option~B is the negative outcome. For Superiority, Option~A indicates that the treatment arm achieved a statistically significant improvement over the comparator arm, and Option~B indicates that it did not. For Endpoint, Option~A indicates that the endpoint was met, or, in the ``at least one arm'' subtype, that at least one arm met it; Option~B indicates that it was not met.

Comparative Effect is a three-way class comparing the treatment (experimental) arm with the comparator arm, which may be a placebo or an active comparator. Option~A indicates that the comparator arm is statistically significantly worse
than the treatment arm. Option~B indicates that the treatment arm is statistically significantly worse than the comparator arm. Option~C indicates that there is no statistically significant difference between the two arms. We include this class because some trials are designed to establish non-inferiority of the experimental treatment relative to the comparator, which is common when the outcome measure concerns safety, tolerability, or the rate of side effects. In these cases a three-way distinction carries more information than a binary one.

Table~\ref{tab:label_dist} reports the label distribution within each class and Table~\ref{tab:benchmark_examples} provides one example question per class.

\begin{table}[ht]
\centering
\small
\setlength{\tabcolsep}{8pt}
\renewcommand{\arraystretch}{1.15}
\begin{tabular}{@{}llrr@{}}
\toprule
\textbf{Question Class} & \textbf{Label} & \textbf{Winter 2025} ($n=605$) & \textbf{Summer 2025} ($n=857$) \\
\midrule
\multirow{2}{*}{Endpoint ($n=80$ / $146$)}
  & Option A & 66 (82.5\%)  & 109 (74.7\%) \\
  & Option B & 14 (17.5\%)  & 37\phn (25.3\%) \\
\midrule
\multirow{2}{*}{Superiority ($n=470$ / $662$)}
  & Option A & 299 (63.6\%) & 400 (60.4\%) \\
  & Option B & 171 (36.4\%) & 262 (39.6\%) \\
\midrule
\multirow{3}{*}{Comparative Effect ($n=55$ / $49$)}
  & Option A & 6\phn (10.9\%)  & 2\phn\phn (4.1\%) \\
  & Option B & 2\phn\phn (3.6\%)   & 1\phn\phn (2.0\%) \\
  & Option C & 47 (85.5\%)  & 46 (93.9\%) \\
\bottomrule
\end{tabular}
\caption{\small Label distribution by question class. Percentages are within class and split. Class sizes are listed as Winter 2025 / Summer 2025 and sum to 605 and 857 respectively. Option~B contains 2 and 1 instances in the Comparative Effect class, so per-class scores on that class are sensitive to individual predictions.}
\label{tab:label_dist}
\end{table}

\begin{table*}[t]
\centering
\caption{Example questions from the benchmark dataset.}
\label{tab:benchmark_examples}
\small
\renewcommand{\arraystretch}{1.3}
\begin{tabularx}{\textwidth}{p{2.8cm} X X X}
\toprule
& \textbf{Superiority Example} & \textbf{Comparative Effect Example} & \textbf{Endpoint Example} \\
\midrule
\textbf{NCT ID} & NCT03335839 & NCT06177912 & NCT05321082 \\
\midrule
\textbf{Outcome Measure} & Post-procedural MBG 0/1 or Distal Embolization & Percentage of participants with solicited systemic AEs & Time to first occurrence of composite endpoint: acute ILD exacerbation, hospitalization for respiratory cause, or death \\
\midrule
\textbf{Outcome Type} & Primary & Primary & Secondary \\
\midrule
\textbf{Time Frame} & 30 days & Up to 5 days & Up to 31 months \\
\midrule
\textbf{Study Arms} &
  \textit{Arm 1:} Intracoronary tPA 10\,mg (Experimental) \newline
  \textit{Arm 2:} Placebo (Placebo Comparator) &
  \textit{Arm 1:} V116 -- Pneumococcal 21-valent conjugate vaccine (Experimental) \newline
  \textit{Arm 2:} PPSV23 -- Pneumococcal 23-valent conjugate vaccine (Active Comparator) &
  \textit{Arm 1:} BI 1015550 low dose / Nerandomilast (Experimental) \newline
  \textit{Arm 2:} BI 1015550 high dose / Nerandomilast (Experimental) \\
\midrule
\textbf{Question} &
  (a) Intracoronary tPA 10\,mg achieved statistically significant improvements over Placebo. \newline
  (b) Intracoronary tPA 10\,mg has \textit{not} achieved statistically significant improvements over Placebo. &
  (a) PPSV23 is statistically significantly worse than V116. \newline
  (b) V116 is statistically significantly worse than PPSV23. \newline
  (c) No statistically significant difference between V116 and PPSV23. &
  (a) At least one arm met this endpoint. \newline
  (b) No arm met this endpoint. \\
\midrule
\textbf{Answer} & \textbf{(b)} & \textbf{(c)} & \textbf{(b)} \\
\bottomrule
\end{tabularx}
\end{table*}

\end{document}